\documentclass[10pt,twocolumn,letterpaper]{article}

\usepackage[pagenumbers]{iccv} %

\def\dsname{VisualPRM400K}

\def\benchmarkname{VisualProcessBench}

\def\modelname{VisualPRM}

\usepackage{booktabs}
\usepackage{multirow}

\usepackage{colortbl}
\usepackage{soul}

\usepackage{marvosym}

\definecolor{mygray}{gray}{.94}
\definecolor{mygreen}{rgb}{0, 0.69, 0.31}

\newcommand\blfootnote[1]{%
\begingroup
\renewcommand\thefootnote{}\footnote{#1}%
\addtocounter{footnote}{-1}%
\endgroup
}

\definecolor{iccvblue}{rgb}{0.21,0.49,0.74}
\usepackage[pagebackref,breaklinks,colorlinks,allcolors=iccvblue]{hyperref}

\def\paperID{1409} %
\def\confName{ICCV}
\def\confYear{2025}

\title{VisualPRM: An Effective Process Reward Model for Multimodal Reasoning}

\author{
\textbf{
    Weiyun Wang$^{1,2}$,
    Zhangwei Gao$^{3,2}$,
    Lianjie Chen$^{4,2}$,
    Zhe Chen$^{5,2}$,
    Jinguo Zhu$^{2}$,
}
\\
\textbf{
    Xiangyu Zhao$^{3,2}$,
    Yangzhou Liu$^{5,2}$,
    Yue Cao$^{5,2}$,
    Shenglong Ye$^{2}$,
    Xizhou Zhu$^{4,2}$,
}
\\
\textbf{
    Lewei Lu$^{7}$,
    Haodong Duan$^{2}$,
    Yu Qiao$^{2}$,
    Jifeng Dai$^{4,2}$,
    Wenhai Wang$^{6,2}$
    \textsuperscript{\Letter}
}
\\
$^1$Fudan University,
$^2$Shanghai AI Laboratory,
\\
$^3$Shanghai Jiaotong University,
$^4$Tsinghua University,
\\
$^5$Nanjing University,
$^6$The Chinese University of Hong Kong,
$^7$SenseTime Research
}

\begin{document}
\maketitle

\blfootnote{{\Letter} Corresponding Author: wangwenhai@pjlab.org.cn}

\begin{abstract}
We introduce {\modelname}, an advanced multimodal Process Reward Model (PRM) with 8B parameters, which improves the reasoning abilities of existing Multimodal Large Language Models (MLLMs) across different model scales and families with Best-of-N (BoN) evaluation strategies.
Specifically, our model improves the reasoning performance of three types of MLLMs and four different model scales. 
Even when applied to the highly capable InternVL2.5-78B, it achieves a 5.9-point improvement across seven multimodal reasoning benchmarks.
Experimental results show that our model exhibits superior performance compared to Outcome Reward Models and Self-Consistency during BoN evaluation.
To facilitate the training of multimodal PRMs, we construct a multimodal process supervision dataset {\dsname} using an automated data pipeline.
For the evaluation of multimodal PRMs, we propose {\benchmarkname}, a benchmark with human-annotated step-wise correctness labels, to measure the abilities of PRMs to detect erroneous steps in multimodal reasoning tasks.
We hope that our work can inspire more future research and contribute to the development of MLLMs.
Our model, data, and benchmark are released in this \href{https://internvl.github.io/blog/2025-03-13-VisualPRM/}{page}.
\end{abstract}

\section{Introduction}
\label{sec:intro}

\begin{figure}[t]
\centering

{\includegraphics[width=\linewidth]{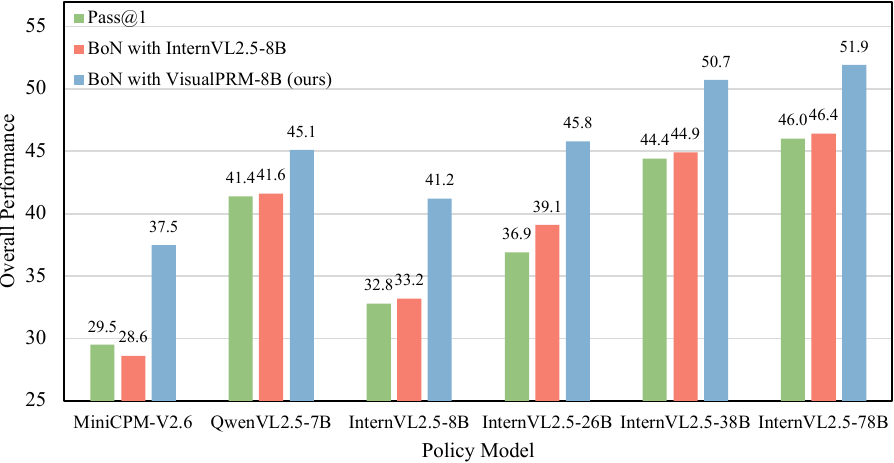}}
\caption{
    \textbf{The overall Best-of-8 evaluation results across seven multimodal reasoning benchmarks with different critic models.}
    Our {\modelname} greatly enhances the overall performance, while InternVL2.5-8B struggles to be an effective critic model.
}
\label{fig:main-teaser}

\vspace{-3mm}

\end{figure}

With the remarkable success of Large Language Models (LLMs) \cite{touvron2023llama,touvron2023llama2,dubey2024llama3,bai2023qwen,2023internlm,cai2024internlm2,brown2020gpt3,openai2023gpt4,claude3series2024} in natural language processing, Multimodal Large Language Models (MLLMs) \cite{wang2023visionllm,li2023blip2,liu2023llava,liu2024llavanext,bai2023qwenvl,wang2023allseeing,wang2024allseeingv2,chen2023internvl,chen2024internvl_1_5,chen2024internvl2_5,gpt4o,reid2024gemini1_5,yao2024minicpm_v} have also achieved significant advancements across various vision-language tasks.
Despite their strong performance in perception and recognition, a large gap remains in reasoning capabilities between open-source and proprietary models.
A series of studies have explored methods to enhance reasoning abilities, focusing on the perspectives of data collection and construction~\cite{muennighoff2025s1,toshniwal2025openmathinstruct,li2024omnicorpus,liu2024mminstruct}, offline preference optimization~\cite{pang2024rpo,wang2024mpo,lai2024stepdpo}, and online reinforcement learning~\cite{shao2024deepseekmath,guo2025deepseek_r1,ahmadian2024rloo,hu2025reinforce++}.
Additionally, another line of research \cite{snell2024llm_tts,dong2024rlhflow,zhang2025qwen_prm,wang2023mathshepherd} investigates utilizing Test-Time Scaling (TTS) to enhance the reasoning abilities of LLMs. This approach requires the policy model to generate multiple response candidates and select the best one, based on the quality estimation of a critic model, thereby improving the response quality at the cost of higher inference time.
However, TTS for MLLMs remains largely unexplored.

This work investigates the application of TTS for MLLMs, focusing on the Best-of-N (BoN) evaluation strategies.
The challenges of adapting TTS for MLLMs involves:
(1) \textit{Lack of effective critic models.}
In BoN evaluation, a critic model is required to estimate the quality of each response candidate.
However, as shown in Figure~\ref{fig:main-teaser}, existing open-source MLLMs struggle to serve as critic models, leading to marginal improvements compared to models without TTS.
This limitation stems from the lack of sufficient critic data in their training corpus.
(2) \textit{Lack of evaluation benchmarks for multimodal critic models.}
The effectiveness of TTS heavily depends on the performance of the critic model. However, directly evaluating critics under BoN settings poses two key issues.
First, the evaluation cost of BoN is expensive.
Although the focus is on the performance of critic models, the policy model is required to generate $N$ reasoning processes, with the majority of computational costs arising from the policy model.
Second, BoN performance is also affected by the policy model, making it difficult to compare different critic models when paired with varying policy models.

To solve these challenges, we first introduce {\dsname}, a dataset comprising approximately 400K multimodal process supervision data.
Each sample includes an image, a question, a step-by-step solution, and correctness annotations for each step.
Specifically, we collect question prompts from MMPR v1.1~\cite{wang2024mpo} and then generate process correctness using an automatic data pipeline~\cite{wang2023mathshepherd}.
This pipeline samples multiple continuations starting from a certain step and computes the expected accuracy of that step as the average accuracy of its continuations.

To facilitate the evaluation of multimodal critic models, we introduce {\benchmarkname}, a benchmark for evaluating PRMs and MLLMs in detecting erroneous steps in multimodal reasoning tasks.
This benchmark includes 2,866 samples with 26,950 human-annotated step-wise correctness labels.
Each sample includes a multimodal reasoning question, a step-by-step solution, and correctness labels for each step.
{To ensure annotation accuracy, we employ human experts with at least a university degree to manually assess the correctness of each step.
Unlike prior benchmarks~\cite{zheng2024processbench,lightman2023prm800k}, which require identifying only the first erroneous step, {\benchmarkname} challenges models to detect all errors within a given solution. This adjustment aligns with recent advancements in model reflection abilities, helping to reduce false negatives in evaluations.
Evaluation results reveal that existing open-source MLLMs struggle to accurately assess step-wise correctness, highlighting the need for improved multimodal critic models.}

Building upon the dataset and benchmark, we develop {\modelname}, an advanced multimodal Process Reward Model (PRM) with 8B parameters, to serve as the critic model in BoN evaluation.
Each training sample is formulated as a multi-turn chat. The first turn includes the image, the question, and the first solution step, while each subsequent turn presents a new step. The model is trained to predict the correctness of the given step at each turn.
\textit{Experimental results demonstrate that \modelname~enhances MLLM reasoning across different model families and scales.}
Specifically, \modelname~improves the overall reasoning performance of MiniCPM-V2.6, QwenVL2.5-7B, InternVL2.5-8B, and InternVL2.5-78B by 8.0, 3.7, 8.4, and 5.9 points, respectively, across seven multimodal reasoning benchmarks~\cite{yue2023mmmu,lu2023mathvista,wang2024mathvision,zhang2024mathverse,zou2024dynamath,qiao2024wemath,xiao2024logicvista}.
Additionally, we compare PRMs with Outcome Reward Models and Self-Consistency in BoN evaluation, finding that PRMs consistently outperform both approaches.

In summary, our main contributions are as follows:

(1) We introduce {\dsname}, a dataset comprising approximately 400K multimodal process supervision data. Building upon this dataset, we develop {\modelname}, an advanced multimodal PRM to serve as the critic model in the BoN evaluation.

(2) We construct {\benchmarkname}, a benchmark designed to measure the abilities of PRMs and MLLMs to identify erroneous steps in multimodal reasoning tasks.
This benchmark comprises 2,866 samples with a total of 26,950 human-annotated step-wise correctness labels.

(3) {Through extensive experiments, we demonstrate that PRMs can serve as effective critic models for test-time scaling of MLLMs.
Specifically, {\modelname} enhances the overall multimodal reasoning performance of MiniCPM-V2.6, QwenVL2.5-7B, InternVL2.5-8B, and InternVL2.5-78B by 8.0, 3.7, 8.4, and 5.9 points, respectively, across seven multimodal reasoning benchmarks.
Furthermore, our results show that PRMs consistently outperform both ORMs and SC in BoN evaluation. Additionally, experiments on {\benchmarkname} reveal that existing open-source MLLMs struggle to accurately assess the correctness of each step.}

\begin{figure*}[t]
\centering

{\includegraphics[width=\linewidth]{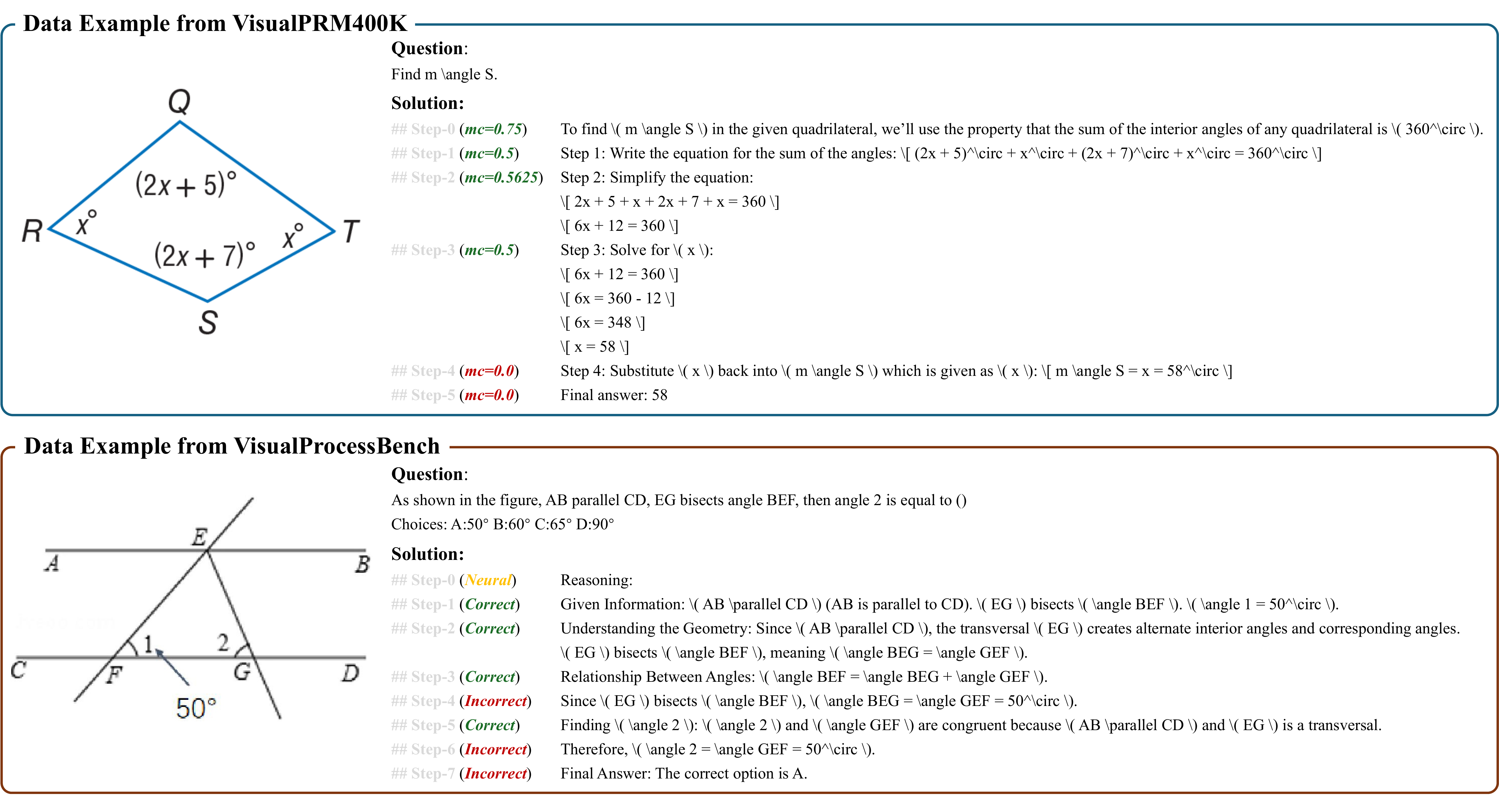}}
\caption{
    \textbf{Data examples in {\dsname} and {\benchmarkname}.}
    For {\dsname}, we generate the data using an automatic data pipeline. The key idea is to estimate the expected accuracy $mc_i$ of the given step $s_{\leq i}$ based on Monte Carlo sampling and consider the step correct if $mc_i>0$. During the training process of {\modelname}, the data is formulated as multi-turn conversations and the model is required to predict the correctness of each step conditioned on the image, question, and previous steps.
    For {\benchmarkname}, we collect questions from existing multimodal reasoning benchmarks~\cite{yue2023mmmu,wang2024mathvision,zhang2024mathverse,zou2024dynamath,qiao2024wemath} and generate the solutions using leading MLLMs~\cite{gpt4o,claude3series2024,chen2024internvl2_5,wang2024mpo,qvq-72b-preview}. Based on these questions and solutions, we employ a team of human experts with at least a university degree to manually annotate the correctness of each step in the solutions.
}
\label{fig:main-data-examples}
\end{figure*}

\section{Related Work}
\label{sec:related_work}

\noindent\textbf{Multimodal Large Language Models.}
A wide range of efforts has been made to advance the development of MLLMs, including improvements in model architecture, data construction, and training algorithms. 
From an architectural perspective, several studies~\cite{liu2023llava,liu2024llavanext,yao2024minicpm_v,chen2024internvl2_5,chen2024internvl_1_5,wang2023allseeing,wang2024allseeingv2,2023interngpt,wang2024mmniah,bai2025qwen2_5_vl,yao2024minicpm_v} employ connectors to align visual embeddings from Vision Foundation Models (VFMs)~\cite{chen2023internvl, zhai2023siglip} with the latent space of LLMs~\cite{bai2023qwen, touvron2023llama, touvron2023llama2, 2023internlm}, achieving promising performance. 
Another series of works~\cite{alayrac2022flamingo,dubey2024llama3,tian2024mminterleaved,wang2023cogvlm} extends pre-trained LLMs with additional layers to fuse visual features, reducing the number of required visual tokens while introducing extra training cost.
In terms of data construction, recent studies have made significant progress~\cite{schuhmann2022laion5b,zhu2024mmc4,laurenccon2024obelics,li2024omnicorpus,liu2024mminstruct,wang2024mpo,zhao2025omnialignv}.
For example, OmniCorpus~\cite{li2024omnicorpus} offers a noisy but large-scale multimodal corpus for pre-training, while MMInstruct~\cite{liu2024mminstruct} provides an open-source, high-quality instruction-tuning dataset. 
Additionally, MMPR~\cite{wang2024mpo} constructs a preference dataset focusing on multimodal reasoning abilities.
Regarding training algorithms, the InternVL2.5 series~\cite{chen2024internvl2_5,wang2024mpo} proposes square loss and Mix Preference Optimization to enhance MLLM capabilities.
Despite these advancements, existing works primarily focus on the training process of MLLMs, leaving Test-Time Scaling (TTS) for MLLMs largely underexplored.
In this work, we investigate TTS applications for MLLMs, specifically focusing on the Best-of-N evaluation to improve multimodal reasoning performance.

\noindent\textbf{Process Reward Models.}
Reward models play a crucial role in Reinforcement Learning (RL)~\cite{schulman2017ppo,shao2024deepseekmath,ahmadian2024rloo,hu2025reinforce++} and TTS~\cite{snell2024llm_tts,wang2023mathshepherd,dong2024rlhflow,luo2024omegaprm}.
Outcome Reward Models (ORMs)~\cite{mcaleese2024openai_critic,zhang2024genrm,ArmoRM} directly assign an overall score to the given response.
In contrast, Process Reward Models (PRMs) first estimate the quality of each step in the given response and then aggregate them into a final score. PRM800K~\cite{lightman2023prm800k} is the first open-source process supervision dataset, entirely annotated by human experts. 
To reduce annotation costs, MathShepherd~\cite{wang2023mathshepherd} and OmegaPRM~\cite{luo2024omegaprm} introduce a Monte Carlo sampling-based data pipeline to automatically estimate the quality of each step.
Despite these advancements in natural language processing, multimodal PRMs remain largely underexplored. In this work, we introduce {\dsname}, the first multimodal process supervision dataset, and develop {\modelname}, a multimodal PRM trained on this dataset.

\noindent\textbf{Benchmarks for Reward Models.}
The evaluation of reward models (RMs) is a crucial research topic.
A series of benchmarks~\cite{lambert2024rewardbench,li2024vlrewardbench,liu2024rm} have been proposed to assess the effectiveness of RMs, typically formulated as a binary preference judgment task. Building on this, subsequent work~\cite{zhou2024rmb} extends the evaluation settings and includes both pairwise and Best-of-N evaluations, providing a more comprehensive evaluation of RM performance.
With the rapid advancement of PRMs, a series of benchmarks~\cite{zheng2024processbench,song2025prmbench} have been introduced to evaluate their step-wise judgment capabilities. Despite these developments, there remains a lack of a multimodal process benchmark. To bridge this gap and support the development of multimodal PRMs, we introduce {\benchmarkname}, a benchmark designed to evaluate the ability of PRMs and MLLMs to detect erroneous steps in multimodal reasoning tasks.

\section{Method}
\label{sec:methods-model}

\begin{figure*}[t]
\centering

{\includegraphics[width=\linewidth]{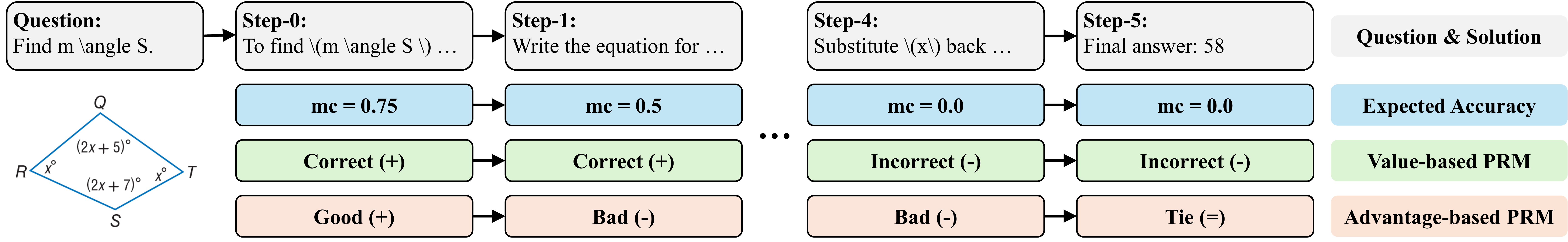}}
\caption{
    \textbf{Different modeling methods for PRMs.}
    PRMs are developed to estimate the quality of each step in a given solution.
    For value-based PRMs, the quality of a certain step is determined by its expected accuracy $mc_i$, where a step is considered correct if $mc_i>0$.
    For advantage-based PRMs, the quality of a certain step is determined by the improvement of $mc_i$ over $mc_{i-1}$, where a step is considered good if $mc_i-mc_{i-1}>0$.
    During the training stage, the output space of PRMs is discretized into specific tokens, while during the inference stage, we compute the step score as the weighted sum of the generation probability for these discretized tokens.
}
\label{fig:main-prm-modeling}

\vspace{-3mm}
\end{figure*}

During Best-of-N (BoN) evaluation, a critic model is required to estimate the quality of each response candidate.
In this work, we formulate the critic model as a Process Reward Model (PRM).
To develop a multimodal PRM, we first construct {\dsname}, a dataset comprising about 400K multimodal process supervision data, 
as detailed in Section~\ref{sec:methods-training-data}. 
We then describe our PRM's modeling approach in Section~\ref{sec:methods-modeling-method}.
Furthermore, to support the evaluation of critic models, we establish {\benchmarkname} to measure the abilities of critic models to detect erroneous steps in multimodal reasoning, as introduced in Section~\ref{sec:methods-benchmark}.

\subsection{{\dsname}}
\label{sec:methods-training-data}

\noindent\textbf{Definition.}
As shown in Figure~\ref{fig:main-data-examples}, each data sample in our {\dsname} consists of an image $I \in \mathcal{I}$, a question $q \in \mathcal{Q}$, a step-by-step solution $s=\{s_0,s_1,\cdots,s_n\} \in \mathcal{S}$, and the expected accuracy annotation $mc=\{mc_0,mc_1,\cdots,mc_n\},mc_i\in\mathbb{R}_{\geq 0}$ for each step, where $n$ is the number of steps of a certain solution and $mc_i$ denotes the expected accuracy of step $s_i$.
The image sets $\mathcal{I}$ and question sets $\mathcal{Q}$ are collected from MMPR v1.1~\cite{wang2024mpo}, while the step-by-step solutions $\mathcal{S}$ are sampled using InternVL2.5 series models~\cite{chen2024internvl2_5,wang2024mpo}.

\noindent\textbf{Process Supervision Generation.}
Given an image $I$, a question $q$, and a solution $s=\{s_0,s_1,\cdots,s_n\}$, we annotate the correctness of each step $s_i$ using an automatic data pipeline.
The key idea is to estimate the expected accuracy of given steps $s_{\leq i}$ based on Monte Carlo sampling.
Specifically, the model is required to complete the solution as follows:
\begin{equation}
    \tilde{s}_{> i} \sim M(\tilde{s}_{> i} \mid I,q, s_{\leq i}),
\end{equation}
where $\tilde{s}_{> i}$ is the completion of $s_{\leq i}$.
Besides, the expected accuracy of $s_i$ is defined as:
\begin{equation}
    mc_i=\frac{\text{num(correct completions)}}{\text{num(sampled completions)}}.
\end{equation}
Notably, to reduce the data construction costs, we set the max number of steps to 12 and evenly merge the steps if the number of current steps exceeds the threshold.

\noindent\textbf{Statistics.}
During the construction process, we sample $4$ solutions for each image-question pair and split each of them into at most $12$ steps. For each step, we sample $16$ continuations and compute $m_i$ according to these continuations.
The resulting dataset comprises approximately {400K} samples and {2 million} steps with process supervision.
Each response averages {126.9} words and {5.6} steps, while each step averages {22.6} words. Among these steps, about {10\%} are incorrect steps.
Despite the imbalanced distribution of correct and incorrect steps, our PRM demonstrates promising performance, as shown in Section~\ref{sec:experiments}.

\subsection{{\modelname}}
\label{sec:methods-modeling-method}

\noindent\textbf{Overview.}
 During the training process, we formulate the process supervision problem as a multi-turn chat task so that we can effectively leverage the generation ability of MLLMs.
The image $I$, question $q$, and the first step $s_0$ of the solution to this question are included in the first turn and a new step is presented in each subsequent turn. The model is required to predict the quality of the given step in each turn as follows:
\begin{equation}
    y_i \sim M(y_i \mid I,q, s_{\leq i}),
\end{equation}
where $y_i$ denotes the quality of $i$-th step.

\noindent\textbf{For value-based PRMs,}
the quality of a certain step is determined by its expected accuracy $mc_i$, which is similar to the definition of the value function in reinforcement learning.
Following Math-Shepherd~\cite {wang2023mathshepherd,dong2024rlhflow}, we require the model to predict the correctness $c_i\in\{+,-\}$ of the given step, rather than the exact score of $mc_i$.
The $i$-th step is considered correct if $mc_i>0$.
We also try to set a threshold to reduce false positive steps, but find that such a threshold negatively impacts the PRM performance, as shown in Section~\ref{sec:suppl-more-ablation}.
Notably, unlike previous works~\cite{wang2023mathshepherd,lightman2023prm800k,dong2024rlhflow}, which choose to supervise only up to the first incorrect step, we always supervise all steps.

\noindent\textbf{For advantage-based PRMs,}
the quality of a certain step is determined by the improvement of $mc_i$ over $mc_{i-1}$, which is analogous to the definition of the advantage function in reinforcement learning.
Similar to value-based PRMs, the quality space is discretized into predefined values $\{+,=,-\}$, meaning that the $i$-th step $s_i$ results in a superior, comparable, or inferior situation.

\noindent\textbf{During inference stage,}
we first compute the scores of each step and then merge them to obtain the response score.
Specifically, the score for each step is defined as the weighted sum of the generation probability for the discretized scores.
For value-based PRMs, the weights for $\{+,-\}$ are $\{1,0\}$.
For advantage-based PRMs, the weights for $\{+,=,-\}$ are $\{1,0,-1\}$.
Without further explanation, we average the scores of each step as the response score.

\subsection{{\benchmarkname}}
\label{sec:methods-benchmark}

\begin{table}[t]

\centering
\small
\renewcommand{\arraystretch}{0.85}

\begin{tabular}{lr}
\toprule
\textbf{Statistics} & \textbf{Number} \\ \midrule
Total Samples       & 2866            \\
\quad - MMMU~\cite{yue2023mmmu}                & 267             \\
\quad - MathVision~\cite{wang2024mathvision}          & 712             \\
\quad - MathVerse~\cite{zhang2024mathverse}           & 1026            \\
\quad - DynaMath~\cite{zou2024dynamath}            & 570             \\
\quad - WeMath~\cite{qiao2024wemath}              & 291             \\ \midrule
Source Solutions    & 2866            \\
\quad - GPT-4o~\cite{gpt4o}              & 870             \\
\quad - Claude-3.5-Sonnet~\cite{claude3series2024}   & 865             \\
\quad - QvQ-72B-Preview~\cite{qvq-72b-preview}     & 825             \\
\quad - InternVL2.5-78B~\cite{chen2024internvl2_5}     & 306             \\ \midrule
Total Steps               & 26950           \\
\quad - Correct Steps       & 16585           \\
\quad - Incorrect Steps     & 7691            \\
\quad - Neural Steps        & 2674            \\ \midrule
Query Word Length Quartile  & (22, 24, 50)             \\
Response Word Length Quartile  & (137, 193, 552)             \\
Step Word Length Quartile  & (13, 31, 67)             \\
Number of Steps per Solution  & 9.4             \\
\bottomrule
\end{tabular}

\caption{
\textbf{Statistics of {\benchmarkname}.}
}
\label{tab:benchmark-statistics}

\vspace{-3mm}

\end{table}

\noindent\textbf{Definition.}
Each sample in our benchmark consists of a multimodal reasoning question, a step-by-step solution, and correctness annotations for each step.
Considering that recent models begin to demonstrate reflection abilities to rectify their own reasoning process, the evaluation setting used in previous works~\cite{zheng2024processbench,lightman2023prm800k}, which only requires the model to find the first erroneous step, may lead to a false negative estimation.
Therefore, our benchmark requires the model to identify all erroneous steps in the given solution instead of only the first erroneous step.

\noindent\textbf{Data Source.}
Our benchmark focuses on multimodal reasoning tasks, collecting images and questions from existing representative multimodal reasoning benchmarks, including MMMU~\cite{yue2023mmmu}, MathVision~\cite{wang2024mathvision}, MathVerse~\cite{zhang2024mathverse}, DynaMath~\cite{zou2024dynamath}, and WeMath~\cite{qiao2024wemath}.
Given these questions, we generate step-by-step solutions using leading MLLMs, including GPT-4o~\cite{gpt4o}, Claude-3.5-Sonnet~\cite{claude3series2024}, Gemini-2.0-Flash~\cite{team2023gemini}, QvQ-72B-Preview~\cite{qvq-72b-preview}, and InternVL2.5-78B~\cite{chen2024internvl2_5}.
The solutions are sampled from different MLLMs to ensure their diversity.

\noindent\textbf{Step Correctness Annotation.}
We employ a team of human experts with at least a university degree to manually annotate the correctness of each step in the solutions. Specifically, 13 people worked for 3 days, resulting in a workload of 39 person-days. The cost per person-day is approximately 37 dollars.
During the annotation process, annotators are provided with the image, question, ground truth answer, and each step of the solution.
Their task is to assign each step in the solution a label of positive, negative, or neutral, as illustrated in Figure~\ref{fig:main-data-examples}. A positive label indicates that the step is correct, while a negative label signifies an incorrect step. The neural label is assigned to steps that do not involve any reasoning process or provide no additional information.
To ensure the annotation quality, annotators are permitted to skip questions they do not understand.
During the annotation process, our dataset is divided into 10 splits, each containing approximately 300 samples. For each split, the authors of this paper manually review about 10\% of the samples. Splits with erroneous annotations are sent back for re-annotation.
See Section~\ref{sec:suppl-benchmark-examples} for more data examples.

\noindent\textbf{Statistics.}
As shown in Table~\ref{tab:benchmark-statistics}, our benchmark comprises 2866 samples.
To enhance the diversity of our evaluation samples, we gather questions and solutions from a wide range of benchmarks and models while carefully regulating their distribution.
The statistics of step distribution are presented in Section~\ref{sec:suppl-benchmark-statistics}.

\noindent\textbf{Metrics.}
In this work, we use macro F1 scores to compare model performance, aiming to mitigate the impact of the imbalanced distribution between correct and incorrect steps.
Specifically, we first compute the F1 scores separately for correct and incorrect steps and then take their average to obtain the overall score.

\section{Experiments}
\label{sec:experiments}

In this section, we first employ {\modelname} to evaluate various MLLMs using BoN evaluation strategies in Section~\ref{sec:experiments-BoN}, demonstrating that PRMs can significantly enhance the reasoning abilities of MLLMs.
Next, we evaluate our {\modelname} and other leading MLLMs on {\benchmarkname} in Section~\ref{sec:experiments-VL-ProcessBench}.
Finally, the ablation studies are presented in Section~\ref{sec:experiments-ablation} and Section~\ref{sec:suppl-more-ablation}.

\begin{table*}[t]

\centering
\small
\setlength\tabcolsep{4.5pt}
\renewcommand{\arraystretch}{0.9}

\begin{tabular}{lcccccccc}
\toprule
\textbf{Model}    & \textbf{MMMU} & \textbf{MathVista} & \textbf{MathVision} & \textbf{MathVerse-VO} & \textbf{DynaMath} & \textbf{WeMath} & \textbf{LogicVista} & \textbf{Overall} \\ \midrule
\multicolumn{9}{c}{\textit{Proprietary Models}}                                                                                                                                                                                                                                                                                                                                    \\ \midrule
GPT-4o~\cite{gpt4o}            & 70.7                                                        & 60.0                                                              & 31.2                                                               & 40.6                                                                     & 34.5              & 45.8            & 52.8                & 47.9             \\
Gemini-2.0-Flash~\cite{reid2024gemini1_5}  & 69.9                                                        & 70.4                                                              & 43.6                                                               & 47.8                                                                     & 42.1              & 47.4            & 52.3                & 53.4             \\
Claude-3.5-Sonnet~\cite{claude3series2024} & 66.4                                                        & 65.3                                                              & 35.6                                                               & 46.3                                                                     & 35.7              & 44.0            & 60.4                & 50.5             \\ \midrule
\multicolumn{9}{c}{\textit{Open-source Models}}                                                                                                                                                                                                                                                                                                                                    \\ \midrule
MiniCPM-V2.6-8B~\cite{yao2024minicpm_v}  & 49.8                                                        & 60.8                                                              & 23.4                                                               & 18.9                                                                     & 9.8               & 16.4            & 27.5                & 29.5             \\
\rowcolor{mygray}
+{\modelname}         & 56.8                                                        & 65.7                                                              & 24.7                                                               & 35.8                                                                     & 11.2              & 31.0            & 37.4                & 37.5             \\
\rowcolor{mygray}
                  & \textcolor{red}{\textbf{+7.0}}                                                & \textcolor{red}{\textbf{+4.9}}                                                      & \textcolor{red}{\textbf{+1.3}}                                                       & \textcolor{red}{\textbf{+16.9}}                                                            & \textcolor{red}{\textbf{+1.4}}      & \textcolor{red}{\textbf{+14.6}}   & \textcolor{red}{\textbf{+9.8}}        & \textcolor{red}{\textbf{+8.0}}     \\ \midrule
Qwen2.5-VL-7B~\cite{bai2025qwen2_5_vl}     & 55.0                                                        & 67.8                                                              & 25.4                                                               & 41.1                                                                     & 21.0              & 35.2            & 44.1                & 41.4             \\
\rowcolor{mygray}
+{\modelname}         & 58.6                                                        & 70.3                                                              & 31.3                                                               & 44.3                                                                     & 23.0              & 39.8            & 48.3                & 45.1             \\
\rowcolor{mygray}
                  & \textcolor{red}{\textbf{+3.6}}                                                & \textcolor{red}{\textbf{+2.5}}                                                      & \textcolor{red}{\textbf{+5.9}}                                                       & \textcolor{red}{\textbf{+3.2}}                                                             & \textcolor{red}{\textbf{+2.0}}      & \textcolor{red}{\textbf{+4.6}}    & \textcolor{red}{\textbf{+4.2}}        & \textcolor{red}{\textbf{+3.7}}     \\ \midrule
InternVL2.5-8B~\cite{chen2024internvl2_5}    & 56.2                                                        & 64.5                                                              & 17.0                                                               & 22.8                                                                     & 9.4               & 23.5            & 36.0                & 32.8             \\
\rowcolor{mygray}
+{\modelname}         & 60.2                                                        & 68.5                                                              & 25.7                                                               & 35.8                                                                     & 18.0              & 36.5            & 43.8                & 41.2             \\
\rowcolor{mygray}
                  & \textcolor{red}{\textbf{+4.0}}                                                & \textcolor{red}{\textbf{+4.0}}                                                      & \textcolor{red}{\textbf{+8.7}}                                                       & \textcolor{red}{\textbf{+13.0}}                                                            & \textcolor{red}{\textbf{+8.6}}      & \textcolor{red}{\textbf{+13.0}}   & \textcolor{red}{\textbf{+7.8}}        & \textcolor{red}{\textbf{+8.4}}     \\ \midrule
InternVL2.5-26B~\cite{chen2024internvl2_5}   & 60.7                                                        & 68.2                                                              & 23.4                                                               & 24.0                                                                     & 11.4              & 30.9            & 39.6                & 36.9             \\
\rowcolor{mygray}
+{\modelname}         & 63.9                                                        & 73.1                                                              & 29.6                                                               & 39.1                                                                     & 23.2              & 40.8            & 51.0                & 45.8             \\
\rowcolor{mygray}
                  & \textcolor{red}{\textbf{+3.2}}                                                & \textcolor{red}{\textbf{+4.9}}                                                      & \textcolor{red}{\textbf{+6.2}}                                                       & \textcolor{red}{\textbf{+15.1}}                                                            & \textcolor{red}{\textbf{+11.8}}     & \textcolor{red}{\textbf{+9.9}}    & \textcolor{red}{\textbf{+11.4}}       & \textcolor{red}{\textbf{+8.9}}     \\ \midrule
InternVL2.5-38B~\cite{chen2024internvl2_5}   & 63.9          & 71.9               & 32.2                & 36.9                  & 20.0              & 38.3            & 47.9                & 44.4             \\
\rowcolor{mygray}
+{\modelname}         & 69.0          & 73.9               & 35.2                & 46.7                  & 30.5              & 46.2            & 53.7                & 50.7             \\
\rowcolor{mygray}
                  & \textcolor{red}{\textbf{+5.1}}                                                & \textcolor{red}{\textbf{+2.0}}                                                      & \textcolor{red}{\textbf{+3.0}}                                                      & \textcolor{red}{\textbf{+9.8}}                                                            & \textcolor{red}{\textbf{+10.5}}     & \textcolor{red}{\textbf{+7.9}}   & \textcolor{red}{\textbf{+5.8}}       & \textcolor{red}{\textbf{+6.3}}    \\ \midrule
InternVL2.5-78B~\cite{chen2024internvl2_5}   & 70.0          & 72.3               & 32.2                & 39.2                  & 19.2              & 39.8            & 49.0                & 46.0             \\
\rowcolor{mygray}
+{\modelname}         & 70.7          & 75.1               & 35.9                & 47.1                  & 31.3              & 49.1            & 53.9                & 51.9             \\
\rowcolor{mygray}
                  & \textcolor{red}{\textbf{+0.7}}                                                & \textcolor{red}{\textbf{+2.8}}                                                      & \textcolor{red}{\textbf{+3.7}}                                                       & \textcolor{red}{\textbf{+7.9}}                                                             & \textcolor{red}{\textbf{+12.1}}     & \textcolor{red}{\textbf{+9.3}}    & \textcolor{red}{\textbf{+4.9}}        & \textcolor{red}{\textbf{+5.9}}     \\ \bottomrule
\end{tabular}

\caption{
\textbf{Results on seven multimodal reasoning benchmarks.}
MMMU~\cite{yue2023mmmu} is a multidisciplinary reasoning benchmark.
MathVista~\cite{lu2023mathvista}, MathVision~\cite{wang2024mathvision}, MathVerse~\cite{zhang2024mathverse}, DynaMath~\cite{zou2024dynamath}, and WeMath~\cite{qiao2024wemath} are mathematics benchmarks. For MathVerse, we report the performance on Vision-Only (VO) split.
LogicVista~\cite{xiao2024logicvista} is a logical reasoning benchmark.
Part of the results are collected from the OpenCompass leaderboard~\cite{opencompass2023}.
The overall score is the average score of the above benchmarks.
By using {\modelname} as the critic model, existing open-source MLLMs achieve significant improvements in reasoning ability under the Best-of-8 evaluation strategy.
}
\label{tab:exp-main-bon}

\end{table*}

\subsection{Results with Best-of-N evaluation}
\label{sec:experiments-BoN}

\noindent\textbf{Benchmarks.}
We evaluate the reasoning abilities of MLLMs across seven benchmarks, including MMMU~\cite{yue2023mmmu}, MathVista~\cite{lu2023mathvista}, MathVision~\cite{wang2024mathvision}, MathVerse~\cite{zhang2024mathverse}, DynaMath~\cite{zou2024dynamath}, WeMath~\cite{qiao2024wemath}, and LogicVista~\cite{xiao2024logicvista}.
The evaluation samples include subject-based, mathematical, and logical reasoning problems.
We report the worst-case accuracy for DynaMath and the overall accuracy for the remaining benchmarks.
For MathVerse, we report the performance on the Vision-Only split.

\noindent\textbf{Settings.}
Without further explanation, we use {\modelname} as the critic model for BoN evaluation and set $N$ to $8$ by default.
The policy model is required to generate $N$ distinct step-by-step Chain-of-Thought (CoT) reasoning processes with a temperature of 0.7.
The response with the highest score is then selected to determine the correctness.

\noindent\textbf{Results.}
As shown in Table~\ref{tab:exp-main-bon}, \textit{{\modelname} greatly enhances the reasoning abilities of MLLMs across different model scales and families.}
Specifically, for models with fewer than 10 billion parameters, the overall performance of InternVL2.5-8B, MiniCPM-V-8B, and Qwen2.5-VL-7B improves by 8.4, 8.0, and 3.7 points, respectively, demonstrating the effectiveness of test-time scaling across different model families.
For larger models, InternVL2.5-26B, InternVL2.5-38B, and InternVL2.5-78B also achieve substantial performance gains over their counterparts without TTS, further validating the scalability and effectiveness of TTS across different model sizes.

\subsection{Results on {\benchmarkname}}
\label{sec:experiments-VL-ProcessBench}

\noindent\textbf{Settings.}
For the evaluation of PRMs, a step is considered correct if the probability of outputting ``$+$'' exceeds that of outputting ``$-$'' by a certain threshold.
For the evaluation of MLLMs, the model is prompted to analyze each step and determine its correctness, classifying it as either correct or incorrect.
When computing the F1 score, we exclude steps labeled as neural by human annotators in Section~\ref{sec:methods-benchmark}.

\noindent\textbf{Results.}
As shown in Table~\ref{tab:exp-main-vlp}, most existing MLLMs struggle to accurately assess the correctness of each step. Specifically, the overall F1 score for random guessing is 50.0, while most open-source MLLMs achieve scores close to this baseline, highlighting their limitations as critic models.
We manually check the judgments of these open-source MLLMs and observe that these models tend to provide positive analysis and label most steps as correct. For example, InternVL2.5-8B achieves an F1 score of 76.8 for positive steps, while its F1 score for negative steps is only 19.2, indicating that InternVL2.5-8B rarely identifies steps as incorrect.
Furthermore, compared to proprietary models, our {\modelname} demonstrates competitive performance, achieving an overall F1 score of 62.0---outperforming GPT-4o and GPT-4o-Mini, and performing on par with Gemini-2.0-Flash. Notably, our model, with only 8 billion parameters, is more efficient than these proprietary counterparts.

\begin{table*}[t]

\centering
\small
\renewcommand{\arraystretch}{0.85}

\begin{tabular}{lcccccc}
\toprule
\textbf{Model}            & \textbf{MMMU} & \textbf{MathVision} & \textbf{MathVerse-VO} & \textbf{DynaMath} & \textbf{WeMath} & \textbf{Overall} \\ \midrule
Random Guessing           & 50.0          & 50.0                & 50.0                  & 50.0              & 50.0            & 50.0             \\ \midrule
\multicolumn{7}{c}{\textit{Proprietary Models}}                                                                                                  \\ \midrule
GPT-4o-Mini~\cite{gpt4o}               & 53.6          & 58.9                & 57.1                  & 56.7              & 58.5            & 57.9             \\
GPT-4o~\cite{gpt4o}                    & 56.3          & 60.2                & 59.7                  & 59.0              & 63.3            & 60.3             \\
Gemini-2.0-Flash~\cite{reid2024gemini1_5}          & 58.5          & 60.1                & 62.8                  & 66.7              & 58.7            & 62.3             \\ \midrule
\multicolumn{7}{c}{\textit{Open-source Models}}                                                                                                  \\ \midrule
MiniCPM-V2.6-8B~\cite{yao2024minicpm_v}          & 44.9          & 50.9                & 58.9                  & 46.7              & 57.4            & 50.4             \\
LLaVA-OV-7B~\cite{li2024llava_onevision}               & 45.7          & 43.0                & 42.2                  & 44.7              & 52.5            & 44.4             \\
LLaVA-OV-72B~\cite{li2024llava_onevision}              & 46.1          & 48.4                & 53.0                  & 57.0              & 57.3            & 52.3             \\
Qwen2.5-VL-7B~\cite{bai2025qwen2_5_vl}             & 53.1          & 51.8                & 47.8                  & 51.3              & 54.2            & 51.0             \\
Qwen2.5-VL-72B~\cite{bai2025qwen2_5_vl}            & 59.2          & 59.0                & 59.7                  & 62.9              & 62.3            & 60.5             \\
InternVL2.5-8B~\cite{chen2024internvl2_5}            & 47.1          & 45.5                & 47.8                  & 50.3              & 50.8            & 48.0             \\
InternVL2.5-26B~\cite{chen2024internvl2_5}           & 48.8          & 47.4                & 49.2                  & 50.4              & 51.4            & 49.2             \\
InternVL2.5-38B~\cite{chen2024internvl2_5}           & 51.5          & 48.4                & 50.9                  & 51.8              & 52.5            & 50.8             \\
InternVL2.5-78B~\cite{chen2024internvl2_5}           & 52.0          & 51.7                & 53.7                  & 50.8              & 52.5            & 52.6             \\ \midrule
\rowcolor{mygray}
{\modelname} (ours) & 58.5          & 62.1                & 61.0                  & 62.7              & 61.8            & 62.0             \\ \bottomrule
\end{tabular}

\caption{
\textbf{Results on {\benchmarkname}.}
We report the macro F1 of the correct and incorrect steps.
The overall score is the micro average of the score from different data sources.
Our {\modelname} exhibits state-of-the-art performance among open-source models.
}
\label{tab:exp-main-vlp}

\vspace{-3mm}

\end{table*}

\subsection{Ablation Studies}
\label{sec:experiments-ablation}

\begin{figure}[t]
\centering
\begin{subfigure}[b]{0.49\linewidth}
    {\includegraphics[width=\textwidth]{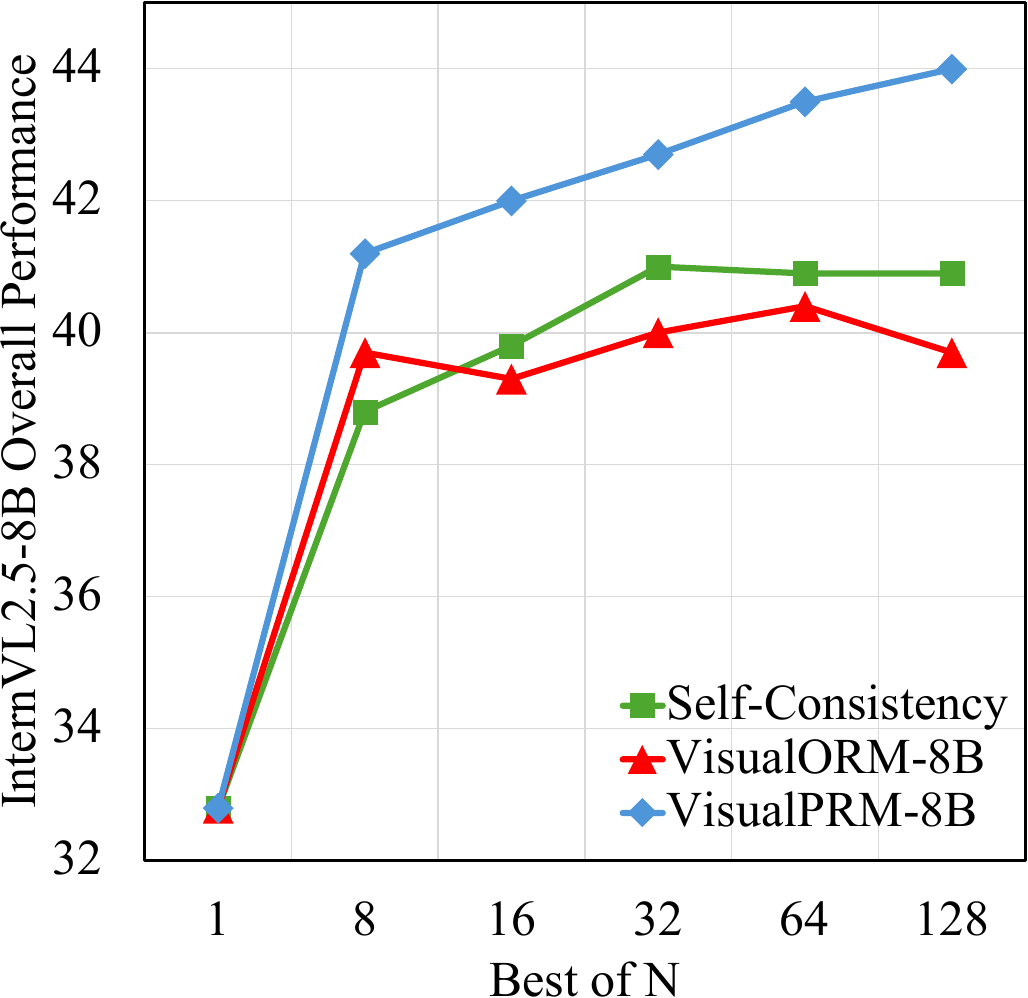}}
    \caption{}
    \label{fig:main-exp-bon-internvl}
\end{subfigure}
\begin{subfigure}[b]{0.49\linewidth}
    {\includegraphics[width=\textwidth]{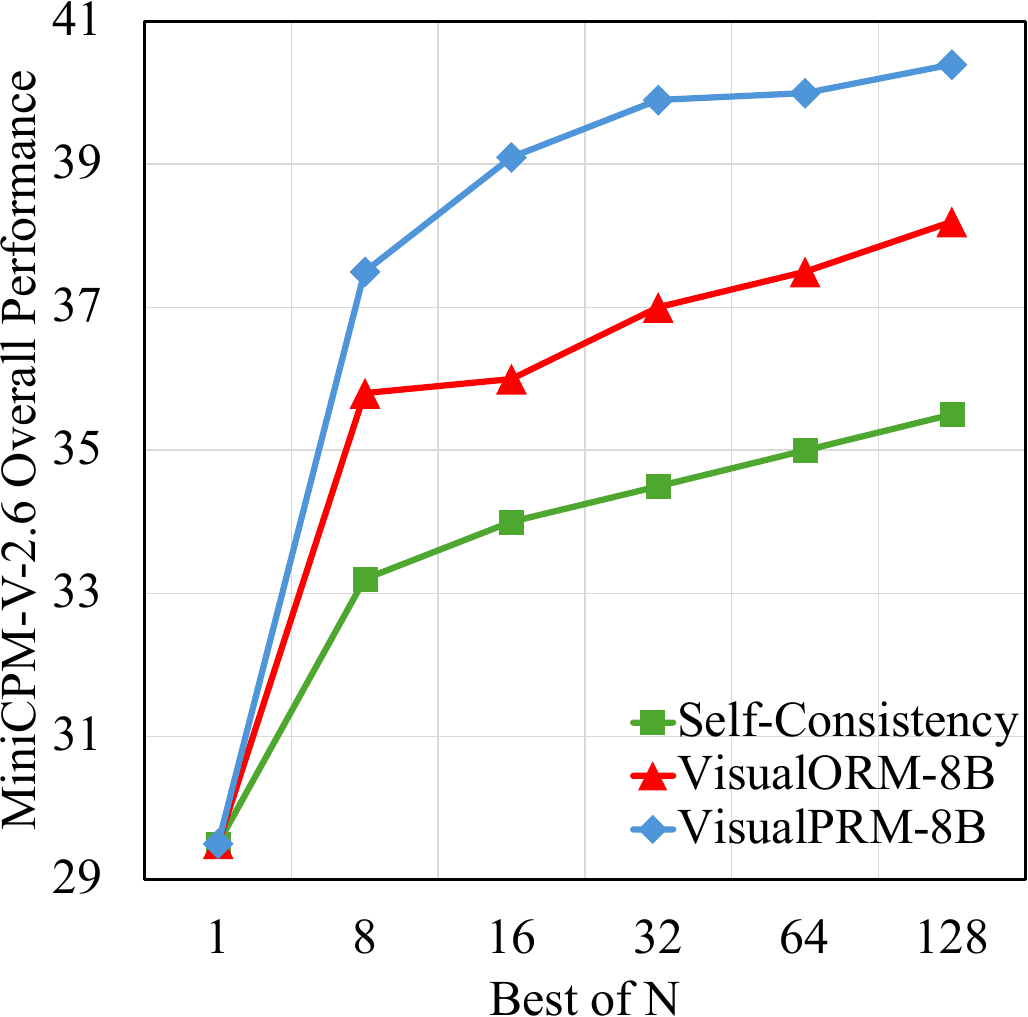}}
    \caption{}
    \label{fig:main-exp-bon-other}
\end{subfigure}
\caption{
    \textbf{Overall Best-of-N results across seven multimodal reasoning benchmarks with different policy and critic models.}
    {\modelname} consistently enhances reasoning performance of both InternVL2.5-8B and MiniCPM-V as $N$ increases and outperforms the improvement introduced by Self-Consistency and ORM, highlighting the effectiveness of PRM in Test-Time Scaling.
}
\label{fig:main-exp-bon}

\vspace{-3mm}

\end{figure}

\noindent\textbf{Effects of BoN.}
Here, we increase the number of response candidates sampled from InternVL2.5-8B and select the final response using Self-Consistency (SC)~\cite{wang2022self_consistency}, Outcome Reward Model (ORM), and PRM.
The training data for ORM are nearly identical to those used for PRM, except that all steps are concatenated into a single step and step-wise correctness annotations are converted into a single correctness label for the outcome.
As shown in Figure~\ref{fig:main-exp-bon}, increasing the number of response candidates $N$ improves the reasoning performance of InternVL2.5-8B and MiniCPM-V2.6-8B when using SC, ORM, or PRM, with PRM yielding the most significant improvements.
Specifically, when using InternVL2.5-8B as the policy model, PRM outperforms SC and ORM by 2.4 and 1.5 points, respectively, under the Best-of-8 evaluation setting.
Moreover, this performance gap widens as $N$ increases, reaching 3.1 and 4.3 points when $N$ is set to $128$.
Notably, when using ORM as the critic model, although performance improves during Best-of-8 evaluation, further increasing $N$ does not lead to consistent gains for InternVL2.5-8B. For example, the Best-of-128 performance is inferior to the Best-of-64 performance.
These results highlight the effectiveness of PRM in TTS.

\noindent\textbf{Effects of PRM modeling methods.}
Here, we compare the value-based PRM and the advantage-based PRM introduced in Section~\ref{sec:methods-modeling-method}, along with different methods for aggregating step scores into a final score, including averaging, as well as selecting the maximum or minimum value.
The results are presented in Table~\ref{tab:exp-ablation-modeling}.
We find that value-based PRMs outperform advantage-based PRMs in both BoN evaluation settings and VL-ProcessBench. We attribute this to the inherent noise in our training data, which is generated through an automatic data pipeline, making it challenging to accurately determine whether a given step contributes to higher or lower expected accuracy.
We also compare two training strategies: supervising all steps (\ie, w/o early stop) versus supervising only up to the first incorrect step (\ie, w. early stop) during training. Experimental results show that the former yields better performance.
Regarding different score aggregation methods, we find that selecting the maximum value results in poorer performance compared to averaging or taking the minimum value.
Analyzing the generated scores reveals that most responses contain a high-scored step, close to $1$, at the beginning of the solution. This phenomenon likely arises because most erroneous steps appear in the middle of the solution.
Our statistics of {\benchmarkname} presented in Section~\ref{sec:suppl-benchmark-statistics} further demonstrate this conclusion.
Furthermore, averaging performs better than selecting the maximum value, likely because the latter relies on a single step's score, while averaging accounts for multiple steps and can be considered as an ensemble approach, which benefits the step quality estimation.

\begin{table}[t]

\centering
\small
\renewcommand{\arraystretch}{0.9}

\begin{tabular}{@{}lcc@{}}
\toprule
\textbf{Critic Model}  & \textbf{BoN} & \textbf{VL-ProcessBench} \\ \midrule
Pass@1                 & 32.8         & -                        \\
Random Guessing        & 33.0         & 50.0                     \\
\midrule
InternVL2.5-8B         & 33.2         & 48.0                     \\
InternVL2.5-78B        & 34.2         & 52.6                     \\
\midrule
Advantage-based PRM    &              &                          \\
+Min                    & 36.8         & 55.0                        \\
+Max                    & 36.9         & 55.0                        \\
+Average                & 37.4         & 55.0                     \\
\midrule
Value (w. early stop)  &              &                          \\
+Min                    & 40.3         & 61.6                        \\
+Max                    & 37.0         & 61.6                        \\
+Average                & 40.6         & 61.6                     \\
\midrule
Value (w/o early stop) &              &                          \\
+Min                    & 40.4         & 62.0                        \\
+Max                    & 35.9         & 62.0                        \\
+Average                & 41.1         & 62.0                     \\ \bottomrule
\end{tabular}

\caption{
\textbf{Comparison of different critic models and score aggregation methods.}
Note that score aggregation methods do not affect performance on {\benchmarkname}, which focuses on step correctness judgement.
We find that supervising all steps (w/o early stop) during training perform slightly better than supervising only up to the first incorrect step (w. early steop).
}
\label{tab:exp-ablation-modeling}

\vspace{-3mm}

\end{table}

\noindent\textbf{MLLM-as-a-Judger.}
Existing MLLMs can be prompted to serve as a critic model. However, as shown in Table~\ref{tab:exp-ablation-modeling}, the InternVL2.5 series struggle to improve BoN performance, resulting in only marginal improvements.
Upon analyzing the generated scores, we find that these models assign similar scores to most solutions.
Consistent with our observations in Section~\ref{sec:experiments-VL-ProcessBench}, the InternVL2.5 series tend to generate positive judgments for most steps, which hinders their ability to effectively distinguish and select the truly superior response.
In addition to their effectiveness as critic models for MLLMs, their inference latency also limits efficiency. Specifically, MLLMs generate judgments for each step in an autoregressive manner, which is time-consuming. In contrast, our {\modelname} computes scores for all steps in a single forward pass by using a “+” as a placeholder for model responses and interpreting its generation probability as the step score.

\noindent\textbf{Results on text-only performance.}
To assess the effectiveness of {\modelname} on text-only inputs, we evaluate the Qwen2.5 series~\cite{yang2024qwen2_5} and InternVL2.5 series~\cite{chen2024internvl2_5} on three text reasoning benchmarks under BoN evaluation settings: GSM8K~\cite{cobbe2021gsm8k}, MATH-500~\cite{hendrycks2021math500}, and GPQA-Diamond~\cite{rein2024gpqa}.
We report accuracy as the evaluation metric for these benchmarks.
As shown in Table~\ref{tab:exp-main-text}, our model enhances the text reasoning abilities of both the Qwen2.5 series and the InternVL2.5 series.
Specifically, Qwen2.5-7B achieves improvements of 6.1 and 5.0 points on MATH-500 and GPQA-Diamond, respectively.
Similarly, Qwen2.5-72B demonstrates gains of 2.1 and 6.6 points on these benchmarks.
For the InternVL2.5 series, InternVL2.5-8B, InternVL2.5-38B, and InternVL2.5-78B achieve improvements of 9.4 and 5.0, 4.6 and 8.1, and 7.4 and 3.5 points, respectively, on MATH-500 and GPQA-Diamond.
These results demonstrate the effectiveness of our {\modelname} in text-only scenarios.

\begin{table}[t]

\centering
\small
\renewcommand{\arraystretch}{0.88}

\begin{tabular}{lccc}
\toprule
\textbf{Model}  & \textbf{GSM8K} & \textbf{MATH-500} & \textbf{GPQA} \\ \midrule
\multicolumn{4}{c}{\textit{Large Language Models}}                           \\ \midrule
Qwen2.5-7B~\cite{yang2024qwen2_5}      & 91.6           & 75.5              & 36.4                  \\
\rowcolor{mygray}
+{\modelname}       & 94.5           & 81.6              & 41.4                  \\
\rowcolor{mygray}
\textbf{}       & \textcolor{red}{\textbf{+2.9}}   & \textcolor{red}{\textbf{+6.1}}      & \textcolor{red}{\textbf{+5.0}}          \\ \midrule
Qwen2.5-32B~\cite{yang2024qwen2_5}     & 95.9           & 83.1              & 49.5                  \\
\rowcolor{mygray}
+{\modelname}       & 96.1           & 85.4              & 53.5                  \\
\rowcolor{mygray}
\textbf{}       & \textcolor{red}{\textbf{+0.2}}   & \textcolor{red}{\textbf{+2.3}}      & \textcolor{red}{\textbf{+4.0}}          \\ \midrule
Qwen2.5-72B~\cite{yang2024qwen2_5}     & 95.8           & 83.1              & 49.0                  \\
\rowcolor{mygray}
+{\modelname}       & 96.5           & 85.2              & 55.6                  \\
\rowcolor{mygray}
\textbf{}       & \textcolor{red}{\textbf{+0.7}}   & \textcolor{red}{\textbf{+2.1}}      & \textcolor{red}{\textbf{+6.6}}          \\ \midrule
\multicolumn{4}{c}{\textit{Multimodal Large Language Models}}                \\ \midrule
InternVL2.5-8B~\cite{chen2024internvl2_5}  & 81.9           & 56.8              & 29.3                  \\
\rowcolor{mygray}
+{\modelname}       & 82.9           & 66.2              & 34.3                  \\
\rowcolor{mygray}
\textbf{}       & \textcolor{red}{\textbf{+1.1}}   & \textcolor{red}{\textbf{+9.4}}      & \textcolor{red}{\textbf{+5.0}}          \\ \midrule
InternVL2.5-38B~\cite{chen2024internvl2_5} & 94.6           & 75.4              & 44.4                  \\
\rowcolor{mygray}
+{\modelname}       & 95.6           & 80.0              & 52.5                  \\
\rowcolor{mygray}
\textbf{}       & \textcolor{red}{\textbf{+1.0}}   & \textcolor{red}{\textbf{+4.6}}      & \textcolor{red}{\textbf{+8.1}}          \\ \midrule
InternVL2.5-78B~\cite{chen2024internvl2_5} & 93.6           & 70.4              & 47.5                  \\
\rowcolor{mygray}
+{\modelname}       & 94.5           & 77.8              & 51.0                  \\
\rowcolor{mygray}
\textbf{}       & \textcolor{red}{\textbf{+0.9}}   & \textcolor{red}{\textbf{+7.4}}      & \textcolor{red}{\textbf{+3.5}}          \\ \bottomrule
\end{tabular}

\caption{
\textbf{Results on text reasoning benchmarks.}
GSM8K and MATH500 are mathematics benchmarks, while GPQA is a multidisciplinary reasoning benchmark.
For GPQA, we report the performance on the Diamond split.
Our {\modelname} enhances the text reasoning abilities of both LLMs and MLLMs under the Best-of-8 evaluation settings.
}
\label{tab:exp-main-text}

\vspace{-3mm}

\end{table}

\section{Conclusion}
\label{sec:conclusion}

In this work, we construct {\dsname}, a dataset comprising about 400K multimodal process supervision data. Building upon this dataset, we develop {\modelname}, an advanced multimodal Process Reward Model (PRM) capable of estimating the value score of each step during the reasoning process.
With the Best-of-N (BoN) evaluation strategies, our model improves the reasoning abilities of existing Multimodal Large Language Models (MLLMs) across different model scales and families.
Experimental results show that our model exhibits superior performance compared to Outcome Reward Models and Self-Consistency during BoN evaluation, highlighting the effectiveness of PRMs in Test-Time Scaling.
To further facilitate the development of multimodal critic models, we construct {\benchmarkname}, a benchmark designed to measure the abilities of PRMs and MLLMs to detect incorrect steps in multimodal reasoning tasks.
Evaluation results show that existing open-source MLLMs struggle to effectively judge the correctness of each step.
We hope that our work can inspire more future research and contribute to the development of MLLMs.

{
    \small
    \bibliographystyle{ieeenat_fullname}
    \bibliography{main}
}

\clearpage
\setcounter{page}{1}
\maketitlesupplementary

\section{Training Hyper-parameters}
\label{sec:suppl-training-hyper}

During the training process of {\modelname}, the data-packing strategy~\cite{chen2024internvl2_5} is enabled during training. We employ the AdamW optimizer~\cite{loshchilov2017adamw} with the $\beta_1$ of $0.9$, the $\beta_2$ of $0.999$, and the weight decay of $0.05$. The learning rate is initialized as $1e\text{-}5$. The training phases include a linear warmup that lasts until the first 5\% of training steps.
The warmup is followed by a cosine decay strategy with a minimum learning rate of 0.
We set the training epoch to 1.

\section{More Ablation Studies}
\label{sec:suppl-more-ablation}

\subsection{Effects of Training Hyper-parameters}

When training our value-based Process Reward Model (PRM) using {\dsname}, we define a step as correct if its expected accuracy exceeds 0.
In this section, we analyze the impact of varying expected accuracy thresholds for determining step correctness. As shown in Table~\ref{tab:suppl-ablation}, increasing the threshold results in a decline in both Best-of-8 evaluation performance and {\benchmarkname} scores.
These results are consistent with the observation in Qwen2.5-Math-PRM~\cite{zhang2025qwen_prm}.
Therefore, we suggest setting the threshold to $0$ during training.

\subsection{Effects of Generation Hyper-parameters}

In this section, we analyze the impact of generation temperature on the Best-of-8 evaluation.
As shown in Table~\ref{tab:suppl-ablation}, as the temperature increases from 0.3 to 1.3, the overall performance of InternVL2.5-8B first improves and then declines. We attribute this phenomenon to the trade-off between response diversity and accuracy. When the temperature is low (\eg, set to 0.3), the generated responses lack diversity, limiting the model's performance upper bound. Conversely, when the temperature is high (\eg, set to 1.3), the responses become more random, reducing the accuracy of individual responses and lowering the model's overall performance ceiling.

\subsection{Effects of Best-of-N evaluation}

In this section, we present the Best-of-N evaluation results as $N$ increases, as shown in Table~\ref{tab:suppl-bon-internvl} and Table~\ref{tab:suppl-bon-minicpm}.
Our results indicate that as $N$ increases, {\modelname} consistently enhances the reasoning abilities of InternVL2.5-8B~\cite{chen2024internvl2_5} and MiniCPM-V2.6~\cite{yao2024minicpm_v}.
Specifically, as $N$ increases from $8$ to $128$, the overall performance of InternVL2.5-8B improves from 41.2 to 44.0, while MiniCPM-V2.6 improves from 37.5 to 40.4, demonstrating the scalability of Test-Time Scaling for MLLMs.

\begin{table*}[t]

\centering
\small
\setlength\tabcolsep{2.5pt}

\begin{tabular}{lr|ccccccc|c}
\toprule
\textbf{Model}                      & \textbf{BoN} & \textbf{MMMU} & \textbf{MathVista} & \textbf{MathVision} & \textbf{MathVerse-VO} & \textbf{DynaMath} & \textbf{WeMath} & \textbf{LogicVista} & \textbf{Overall} \\ \midrule
\multirow{6}{*}{Self Consistency}    & 1            & 56.2          & 64.5               & 17.0                & 22.8                  & 9.4               & 23.5            & 36.0                & 32.8             \\
                                    & 8            & 58.0          & 65.9               & 23.4                & 30.5                  & 18.4              & 32.7            & 43.0                & 38.8             \\
                                    & 16           & 58.6          & 65.8               & 26.3                & 32.1                  & 19.4              & 33.0            & 43.4                & 39.8             \\
                                    & 32           & 60.4          & 66.7               & 28.0                & 32.6                  & 20.8              & 34.1            & 44.7                & 41.0             \\
                                    & 64           & 59.7          & 66.7               & 26.6                & 33.2                  & 20.6              & 35.8            & 43.4                & 40.9             \\
                                    & 128          & 60.6          & 67.4               & 25.7                & 32.0                  & 22.6              & 34.7            & 43.2                & 40.9             \\ \midrule
\multirow{6}{*}{VisualORM} & 1            & 56.2          & 64.5               & 17.0                & 22.8                  & 9.4               & 23.5            & 36.0                & 32.8             \\
                                    & 8            & 60.2          & 67.0               & 25.3                & 32.5                  & 16.4              & 35.0            & 41.8                & 39.7             \\
                                    & 16           & 58.3          & 67.7               & 27.0                & 33.6                  & 16.6              & 33.1            & 39.1                & 39.3             \\
                                    & 32           & 58.6          & 67.9               & 26.3                & 33.6                  & 17.4              & 34.4            & 42.1                & 40.0             \\
                                    & 64           & 59.4          & 66.8               & 28.6                & 33.9                  & 17.8              & 34.1            & 42.3                & 40.4             \\
                                    & 128          & 59.4          & 66.6               & 28.3                & 33.5                  & 16.8              & 32.3            & 40.9                & 39.7             \\ \midrule
\multirow{6}{*}{{\modelname}} & 1            & 56.2          & 64.5               & 17.0                & 22.8                  & 9.4               & 23.5            & 36.0                & 32.8             \\
                                    & 8            & 60.2          & 68.5               & 25.7                & 35.8                  & 18.0              & 36.5            & 43.8                & 41.2             \\
                                    & 16           & 60.2          & 69.9               & 27.3                & 36.4                  & 19.0              & 38.8            & 42.5                & 42.0             \\
                                    & 32           & 60.3          & 70.4               & 29.6                & 37.8                  & 17.2              & 40.3            & 43.4                & 42.7             \\
                                    & 64           & 61.4          & 69.6               & 30.6                & 38.2                  & 18.8              & 40.2            & 45.4                & 43.5             \\
                                    & 128          & 61.7          & 70.8               & 30.3                & 39.3                  & 19.4              & 40.9            & 45.4                & 44.0             \\ \bottomrule
\end{tabular}

\caption{
\textbf{Overall Best-of-N results of InternVL2.5-8B across seven multimodal reasoning benchmarks with different critic models.}
}
\label{tab:suppl-bon-internvl}

\end{table*}

\begin{table*}[t]

\centering
\small
\setlength\tabcolsep{2.5pt}

\begin{tabular}{lr|ccccccc|c}
\toprule
\textbf{Model}                      & \textbf{BoN} & \textbf{MMMU} & \textbf{MathVista} & \textbf{MathVision} & \textbf{MathVerse-VO} & \textbf{DynaMath} & \textbf{WeMath} & \textbf{LogicVista} & \textbf{Overall} \\ \midrule
\multirow{6}{*}{Self Consistency}    & 1            & 49.8          & 60.8               & 23.4                & 18.9                  & 9.8               & 16.4            & 27.5                & 29.5             \\
                                    & 8            & 51.8          & 58.9               & 21.7                & 31.5                  & 10.0              & 22.6            & 35.6                & 33.2             \\
                                    & 16           & 51.7          & 60.2               & 21.7                & 31.5                  & 11.6              & 25.7            & 35.3                & 34.0             \\
                                    & 32           & 52.2          & 60.1               & 24.3                & 33.1                  & 11.4              & 24.3            & 36.0                & 34.5             \\
                                    & 64           & 51.7          & 61.0               & 23.4                & 34.8                  & 12.8              & 25.8            & 35.3                & 35.0             \\
                                    & 128          & 53.2          & 61.7               & 25.7                & 33.5                  & 13.0              & 25.6            & 35.6                & 35.5             \\ \midrule
\multirow{6}{*}{VisualORM} & 1            & 49.8          & 60.8               & 23.4                & 18.9                  & 9.8               & 16.4            & 27.5                & 29.5             \\
                                    & 8            & 55.7          & 66.0               & 22.0                & 33.5                  & 10.2              & 24.1            & 38.9                & 35.8             \\
                                    & 16           & 56.4          & 65.3               & 24.0                & 32.1                  & 10.4              & 27.3            & 36.5                & 36.0             \\
                                    & 32           & 58.8          & 64.8               & 19.7                & 35.7                  & 12.0              & 29.4            & 38.5                & 37.0             \\
                                    & 64           & 58.2          & 67.3               & 22.7                & 35.5                  & 11.0              & 30.1            & 37.6                & 37.5             \\
                                    & 128          & 58.2          & 66.5               & 25.3                & 35.4                  & 11.6              & 30.0            & 40.7                & 38.2             \\ \midrule
\multirow{6}{*}{{\modelname}} & 1            & 49.8          & 60.8               & 23.4                & 18.9                  & 9.8               & 16.4            & 27.5                & 29.5             \\
                                    & 8            & 56.8          & 65.7               & 24.7                & 35.8                  & 11.2              & 31.0            & 37.4                & 37.5             \\
                                    & 16           & 58.8          & 68.6               & 24.0                & 37.3                  & 12.4              & 32.7            & 39.8                & 39.1             \\
                                    & 32           & 57.8          & 68.4               & 26.6                & 38.5                  & 13.4              & 35.3            & 39.1                & 39.9             \\
                                    & 64           & 58.6          & 69.4               & 25.3                & 39.7                  & 12.2              & 38.2            & 36.9                & 40.0             \\
                                    & 128          & 59.3          & 69.4               & 25.3                & 39.1                  & 14.4              & 37.0            & 38.3                & 40.4             \\ \bottomrule
\end{tabular}

\caption{
\textbf{Overall Best-of-N results of MiniCPM-V2.6 across seven multimodal reasoning benchmarks with different critic models.}
}
\label{tab:suppl-bon-minicpm}

\end{table*}

\begin{table*}[t]

\centering
\small
\setlength\tabcolsep{2pt}

\begin{tabular}{lccccccccc}
\toprule
\textbf{Model}  & \textbf{MMMU} & \textbf{MathVista} & \textbf{MathVision} & \textbf{MathVerse-VO} & \textbf{DynaMath} & \textbf{WeMath} & \textbf{LogicVista} & \textbf{Overall} & \textbf{VL-ProcessBench} \\ \midrule
\multicolumn{10}{c}{\textit{Threshold}}                                                                                                                                                                      \\ \midrule
Threshold=0.00  & 59.3          & 68.5               & 25.7                & 35.8                  & 18.0              & 36.5            & 43.8                & 41.1             & 62.0                     \\
Threshold=0.625 & 59.7          & 66.8               & 24.7                & 36.7                  & 18.4              & 35.0            & 41.8                & 40.4             & 61.0                     \\
Threshold=0.125 & 58.0          & 67.9               & 27.6                & 35.4                  & 17.4              & 35.3            & 41.6                & 40.5             & 60.7                     \\
Threshold=0.25  & 58.6          & 67.6               & 25.7                & 33.6                  & 16.8              & 36.0            & 41.4                & 40.0             & 60.2                     \\ \midrule
\multicolumn{10}{c}{\textit{Temperature}}                                                                                                                                                                    \\ \midrule
Temperature=0.3 & 59.7          & 69.4               & 26.0                & 32.6                  & 17.6              & 35.5            & 42.7                & 40.5             & -                        \\
Temperature=0.7 & 59.3          & 68.5               & 25.7                & 35.8                  & 18.0              & 36.5            & 43.8                & 41.1             & -                        \\
Temperature=1.0 & 61.7          & 67.2               & 27.3                & 35.8                  & 16.6              & 34.2            & 43.2                & 40.9             & -                        \\
Temperature=1.3 & 57.9          & 66.1               & 25.0                & 32.1                  & 16.8              & 31.9            & 40.5                & 38.6             & -                        \\ \bottomrule
\end{tabular}

\caption{
\textbf{Ablation studies about the effects of expected accuracy threshold and generationo temperaure.}
}
\label{tab:suppl-ablation}

\end{table*}

\section{More Statistics for {\benchmarkname}}
\label{sec:suppl-benchmark-statistics}

\begin{figure}[t!]
\centering
\begin{subfigure}[b]{0.47\linewidth}
    {\includegraphics[width=\textwidth]{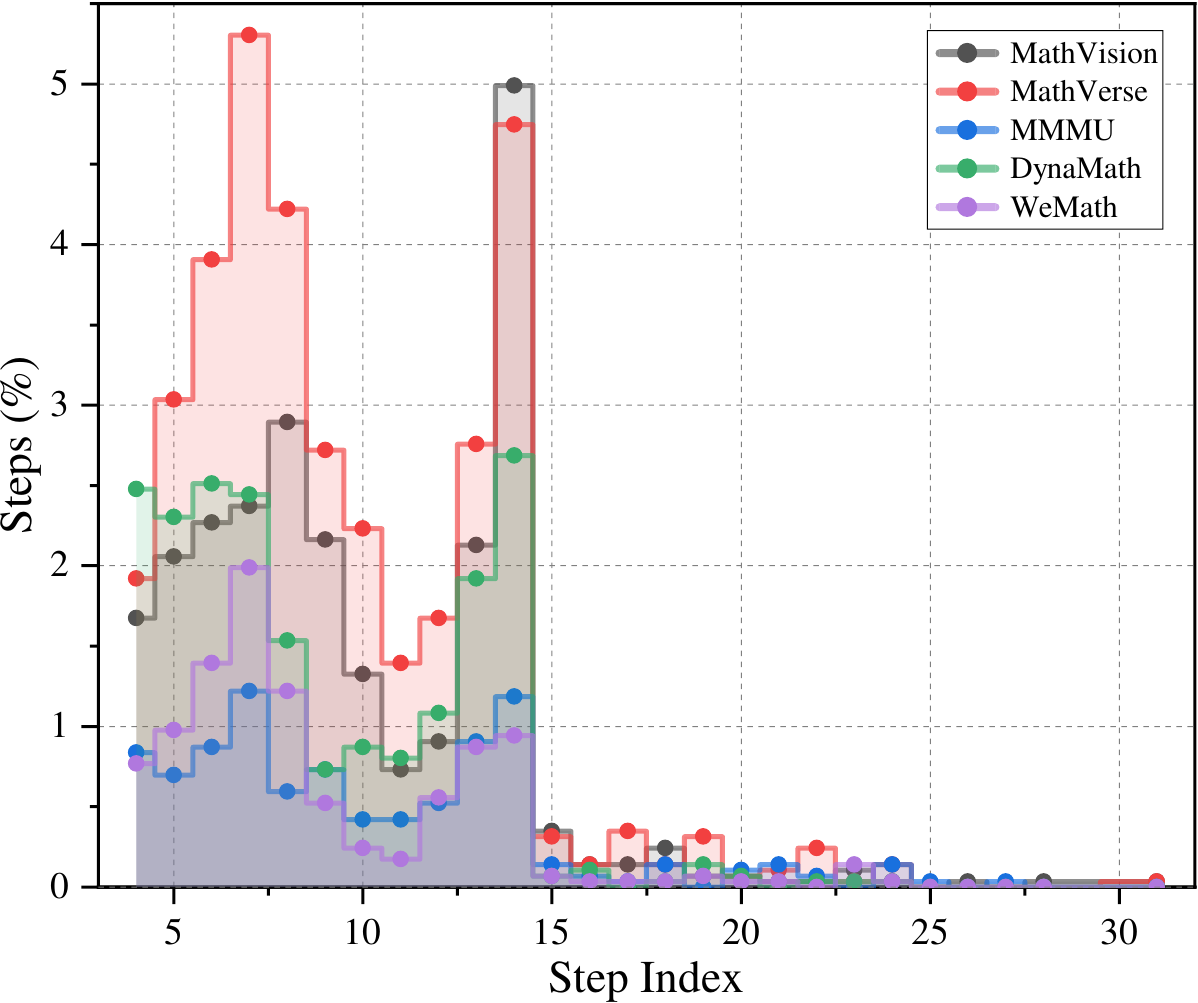}}
    \caption{}
    \label{fig:suppl-dist-num-steps}
\end{subfigure}
\begin{subfigure}[b]{0.47\linewidth}
    {\includegraphics[width=\textwidth]{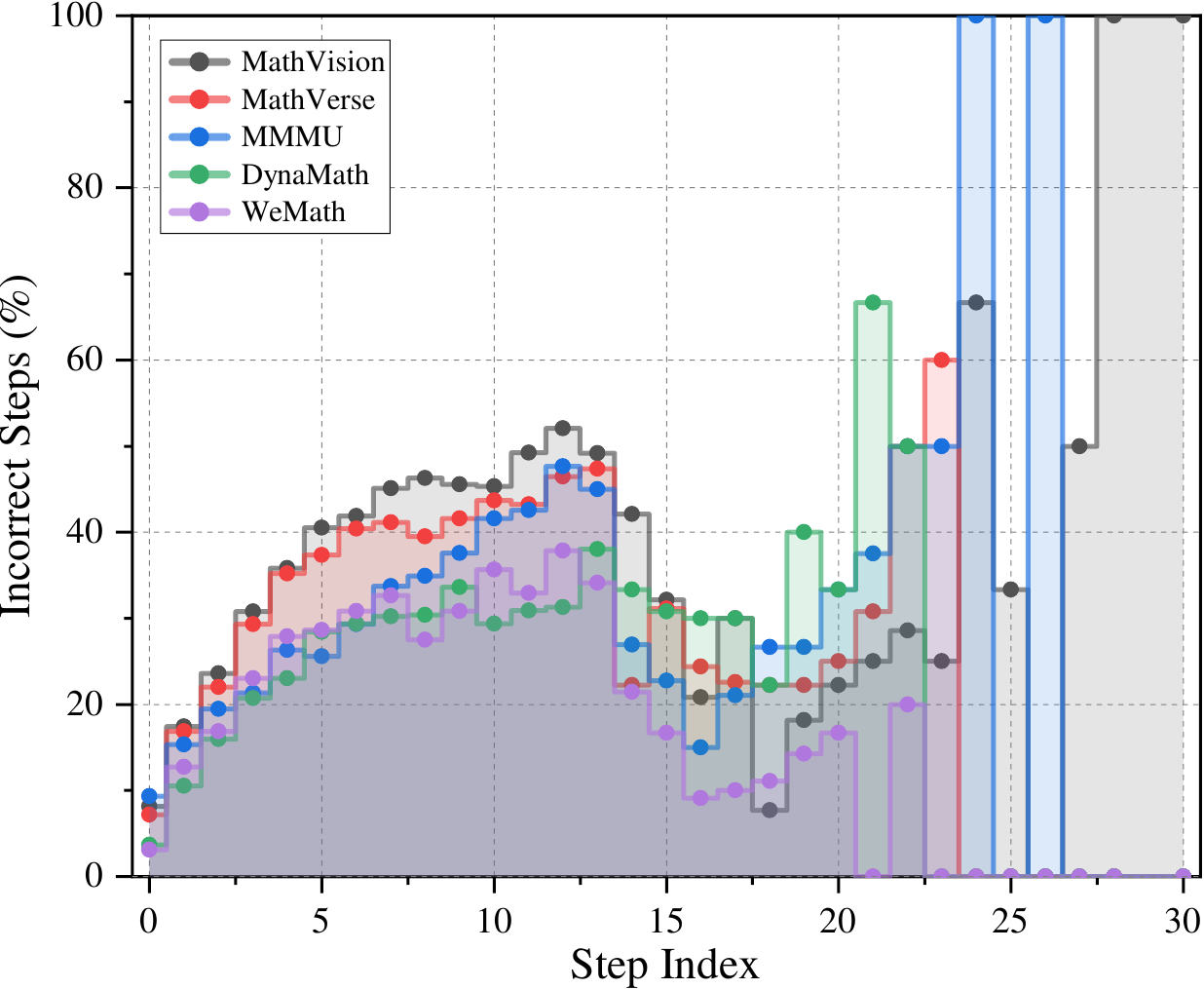}}
    \caption{}
    \label{fig:suppl-dist-incorrect-steps}
\end{subfigure}
\caption{
\textbf{Step Distribution of {\benchmarkname}.}
The X-axis represents the step index.
(a) The Y-axis indicates the proportion of steps at each index relative to the total number of steps, reflecting the distribution of step positions in solutions.
(b) The Y-axis represents the error rate of steps at each index, showing the likelihood of errors occurring at different step positions.
}
\label{fig:suppl-benchmark-dist}

\vspace{-2mm}
\end{figure}

The statistics for step distribution of {\benchmarkname} is presented in Figure~\ref{fig:suppl-benchmark-dist}.
We observe that most solutions consist of fewer than 15 steps. Among these solutions with fewer than 15 steps, most solutions contain about {7} or {13} steps.
For the correctness of each step, we observe that the error rate is lower in the first three steps and then increases as the step index grows. We attribute this to the fact that problems requiring more reasoning steps tend to be more challenging, leading to a gradual rise in step error rates. Notably, starting from step 15, the error rate drops sharply. This is because the number of steps in this range is relatively small, resulting in significant statistical fluctuations.

\section{More Data Examples in {\dsname}}
\label{sec:suppl-training-data-examples}
In this section, we provide more data examples of {\dsname} in Figure~\ref{fig:suppl-data-examples} from different domains, including general visual question answering (VQA)~\cite{goyal2017vqav2,hudson2019gqa,marino2019okvqa,lu2021iconqa}, science~\cite{kembhavi2016ai2d,chen2024m3cot,lu2022scienceqa}, chart~\cite{masry2022chartqa,kafle2018dvqa,chang2022mapqa},  mathematics~\cite{lindstrom2022clevrmath,seo2015geos,cao2022geoqa_plus,lu2021geometry3k,kazemi2023geomverse,gao2023gllava}, OCR~\cite{mathew2022infographicvqa,singh2019textvqa,biten2019stvqa,huang2019sroie,mishra2019ocrvqa}, and document~\cite{clark2017docqa}.

\section{More Data Examples in {\benchmarkname}}
\label{sec:suppl-benchmark-examples}
In this section, we provide more data examples in {\benchmarkname} from different data sources.
Specifically, we randomly choose three examples from our benchmark and visualize them in Figure~\ref{fig:suppl-benchmark-examples}.
Additionally, in Figure~\ref{fig:suppl-benchmark-reflection}, we provide an example where the model initially generates an incorrect answer and then autonomously corrects it.

\begin{figure*}[t]
\centering
\begin{subfigure}[b]{\textwidth}
    {\includegraphics[width=\textwidth]{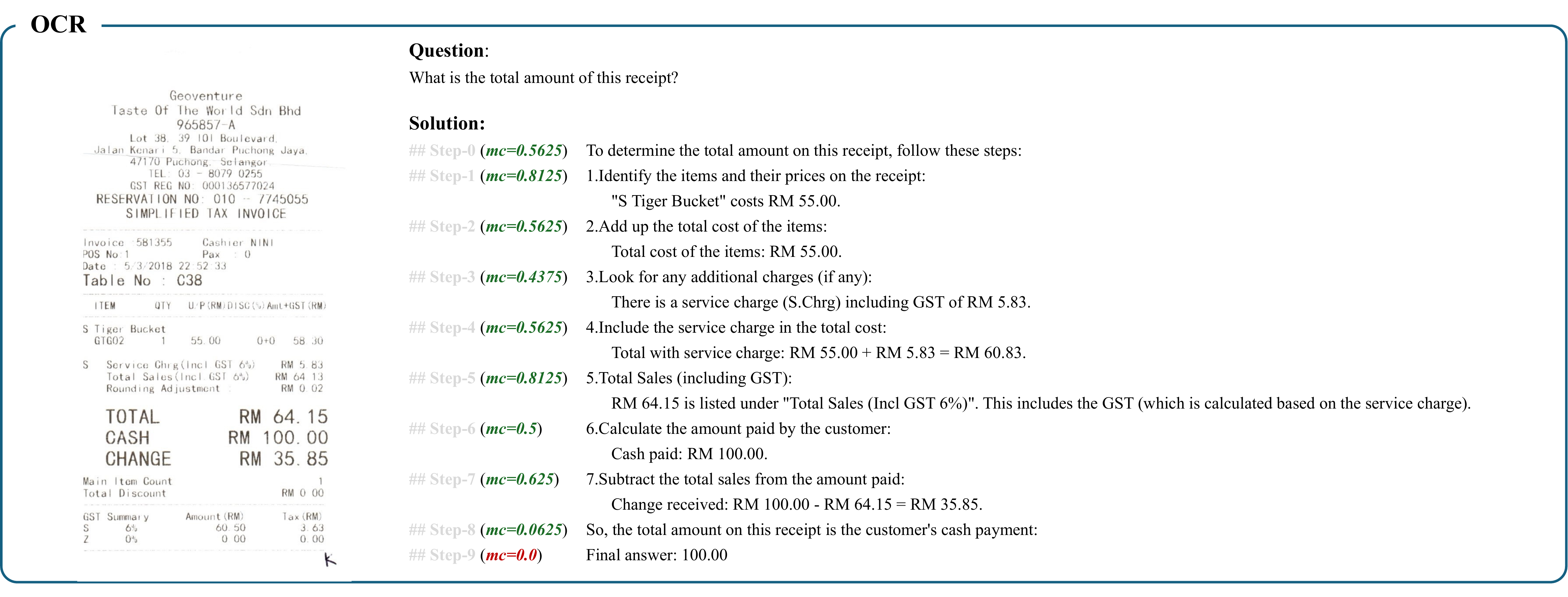}}
    \caption{}
\end{subfigure}
\begin{subfigure}[b]{\textwidth}
    {\includegraphics[width=\textwidth]{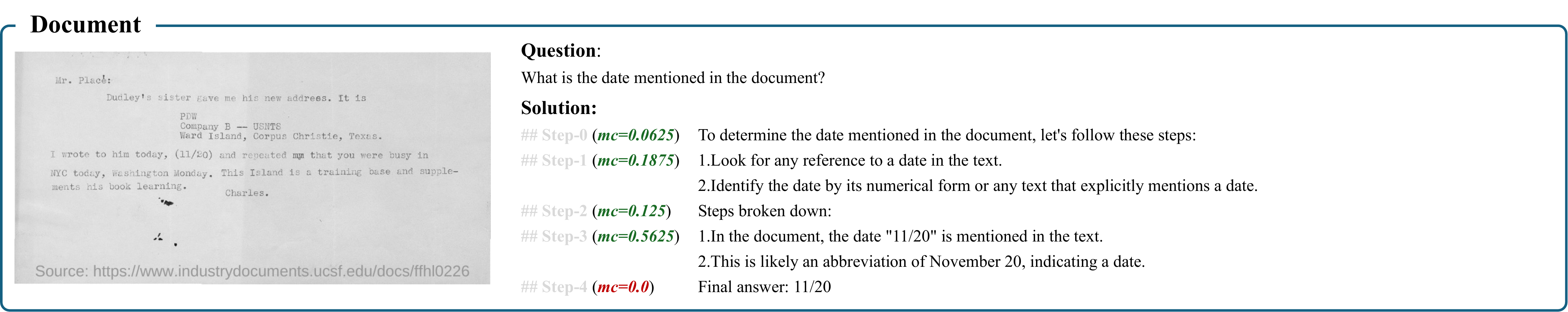}}
    \caption{}
\end{subfigure}
\end{figure*}

\addtocounter{figure}{-1}

\begin{figure*}[t]
\centering
\begin{subfigure}[b]{\textwidth}
    \setcounter{subfigure}{2} %
    {\includegraphics[width=\textwidth]{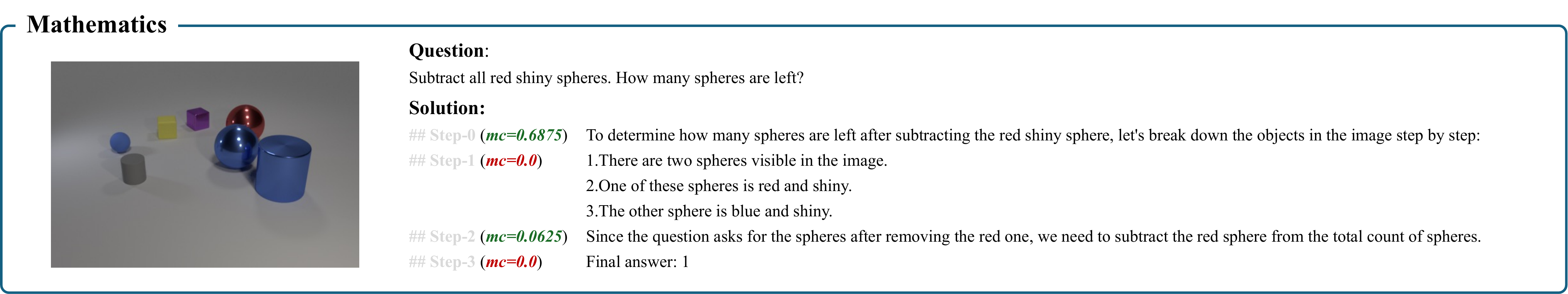}}
    \caption{}
\end{subfigure}
\begin{subfigure}[b]{\textwidth}
    {\includegraphics[width=\textwidth]{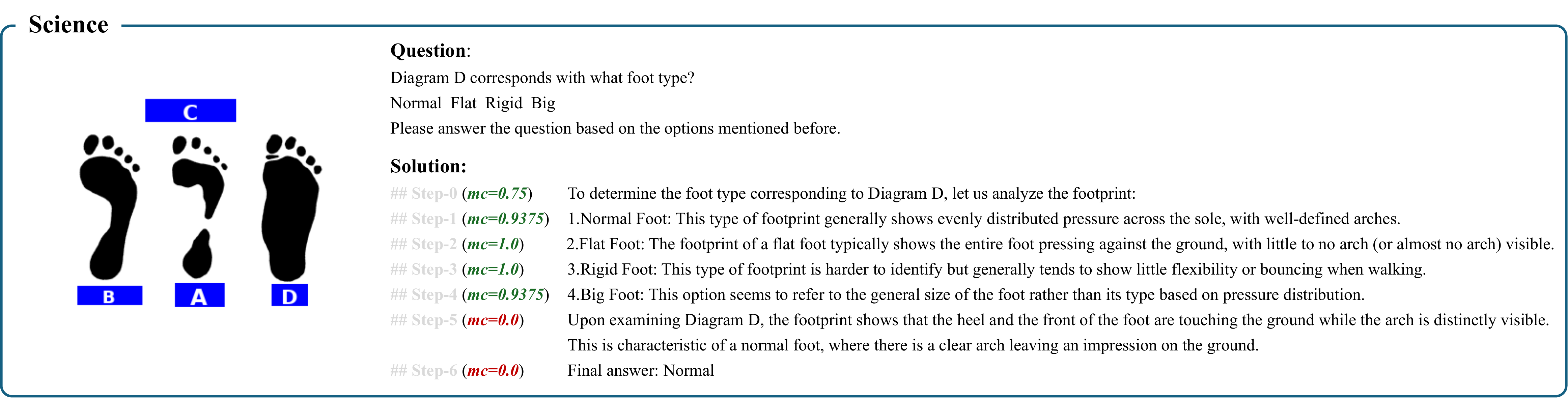}}
    \caption{}
\end{subfigure}
\begin{subfigure}[b]{\textwidth}
    {\includegraphics[width=\textwidth]{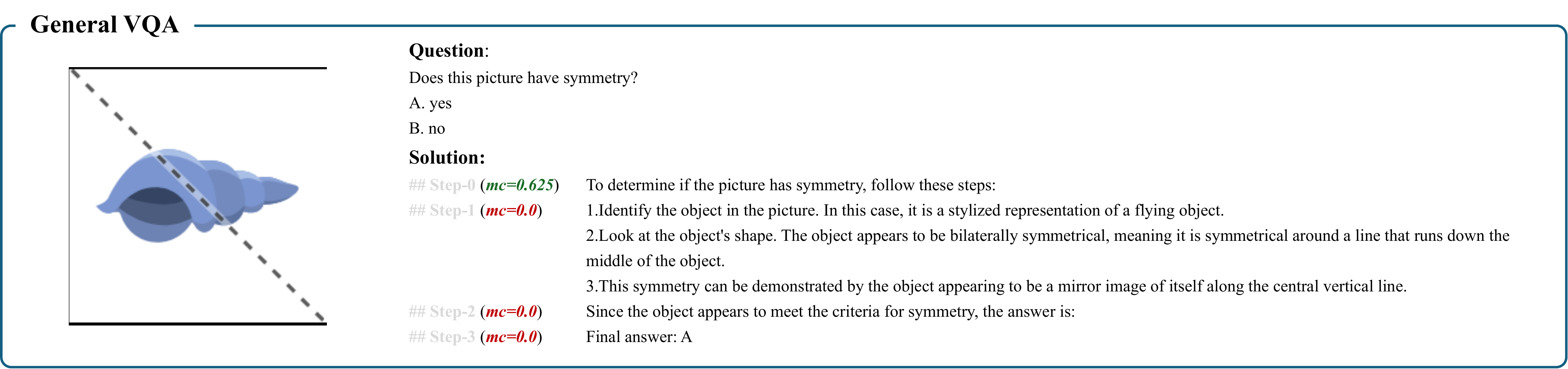}}
    \caption{}
\end{subfigure}
\begin{subfigure}[b]{\textwidth}
    {\includegraphics[width=\textwidth]{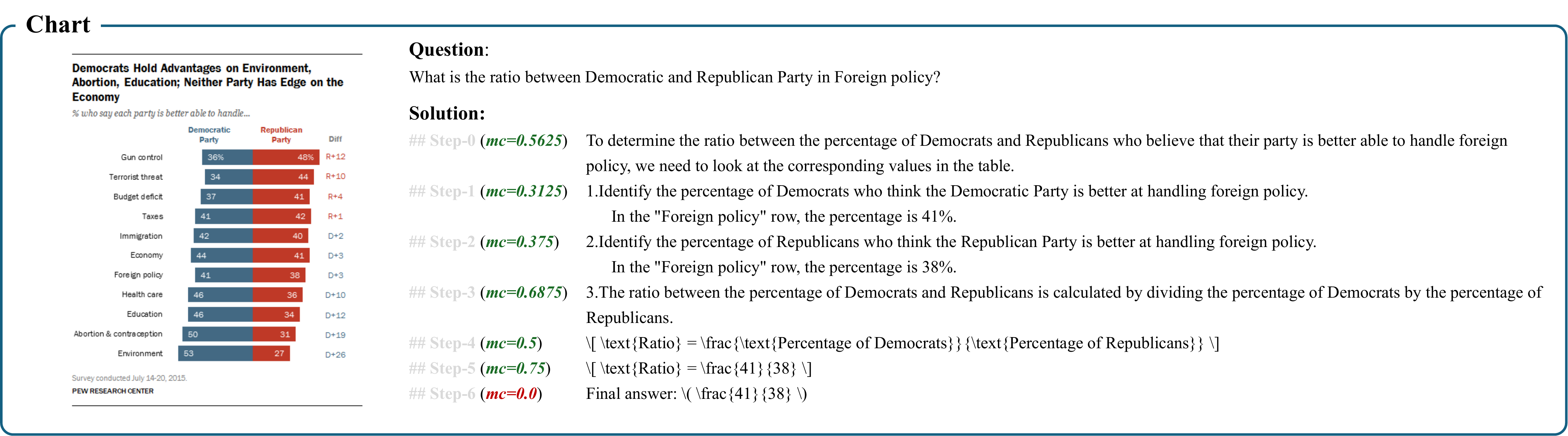}}
    \caption{}
\end{subfigure}
\caption{
\textbf{More data examples from {\dsname}.}
}
\label{fig:suppl-data-examples}
\end{figure*}

\begin{figure*}[t]
\centering
\begin{subfigure}[b]{\textwidth}
    {\includegraphics[width=\textwidth]{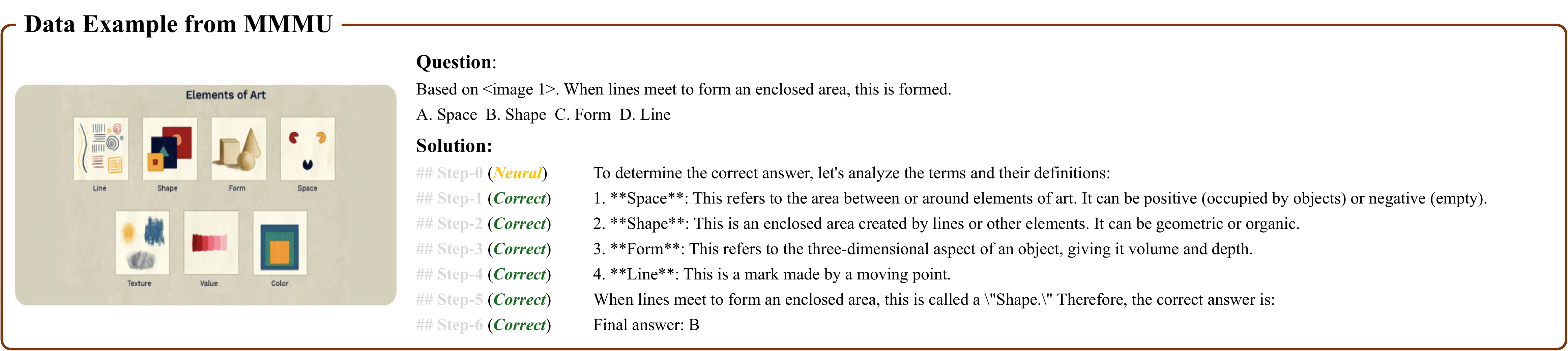}}
    \caption{}
\end{subfigure}
\begin{subfigure}[b]{\textwidth}
    {\includegraphics[width=\textwidth]{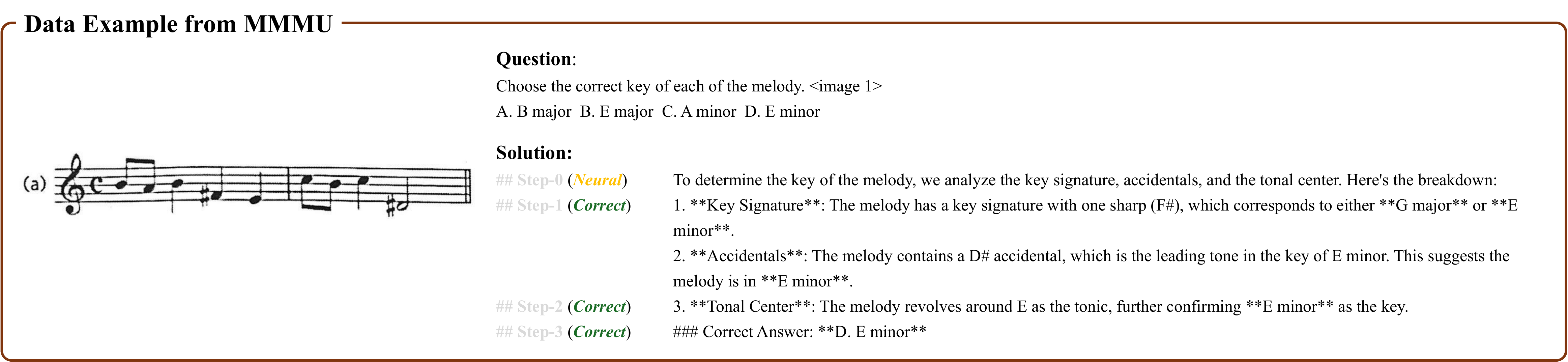}}
    \caption{}
\end{subfigure}
\begin{subfigure}[b]{\textwidth}
    {\includegraphics[width=\textwidth]{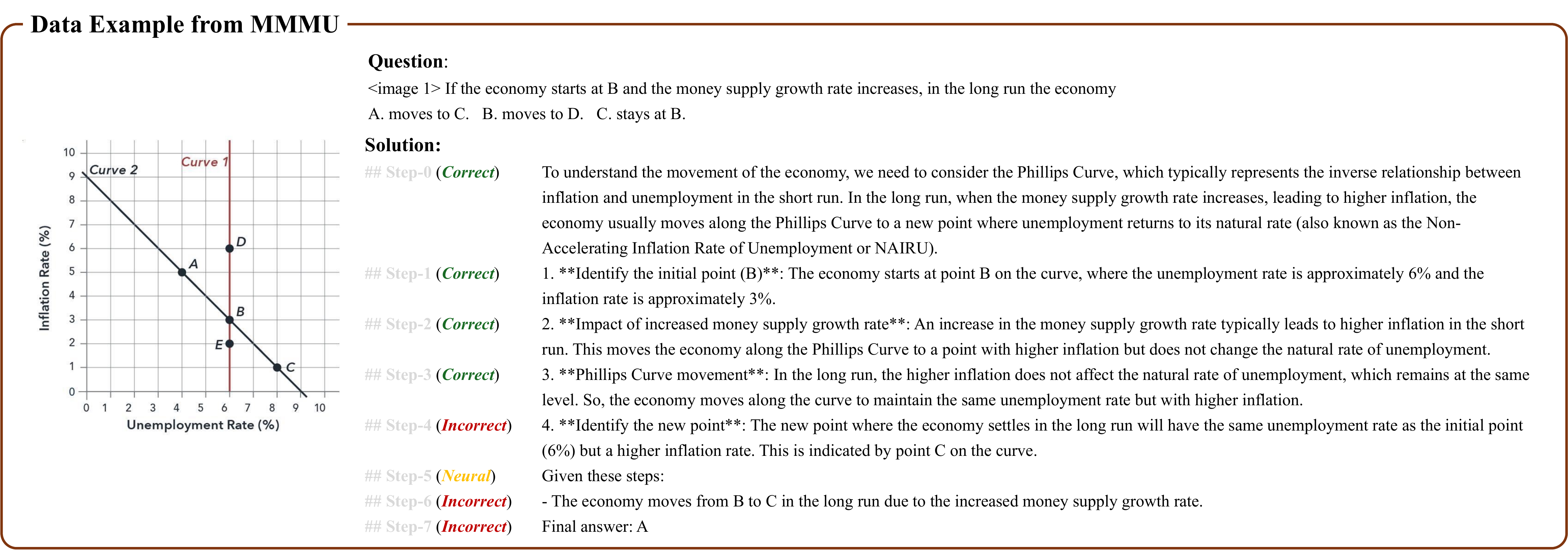}}
    \caption{}
\end{subfigure}
\begin{subfigure}[b]{\textwidth}
    {\includegraphics[width=\textwidth]{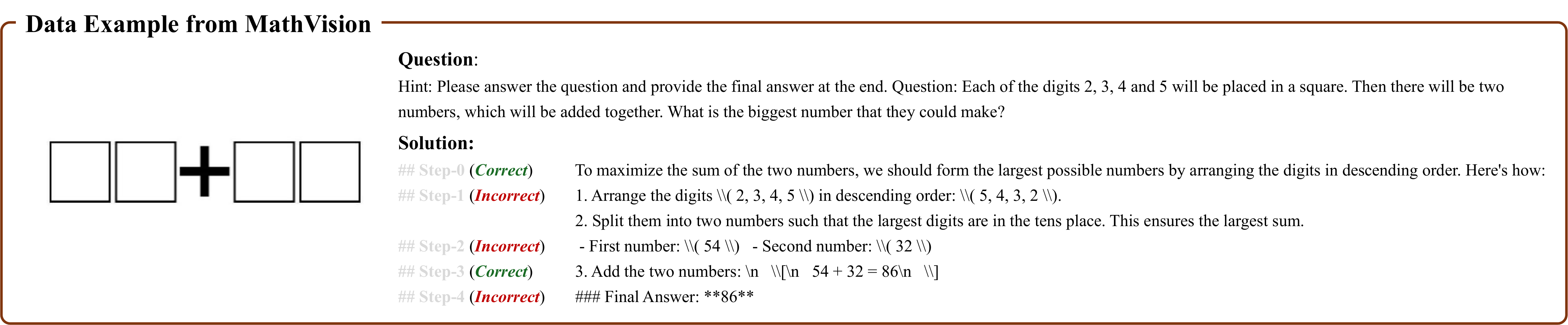}}
    \caption{}
\end{subfigure}
\end{figure*}

\addtocounter{figure}{-1}

\begin{figure*}[t]
\centering
\begin{subfigure}[b]{\textwidth}
    \setcounter{subfigure}{4} %
    {\includegraphics[width=\textwidth]{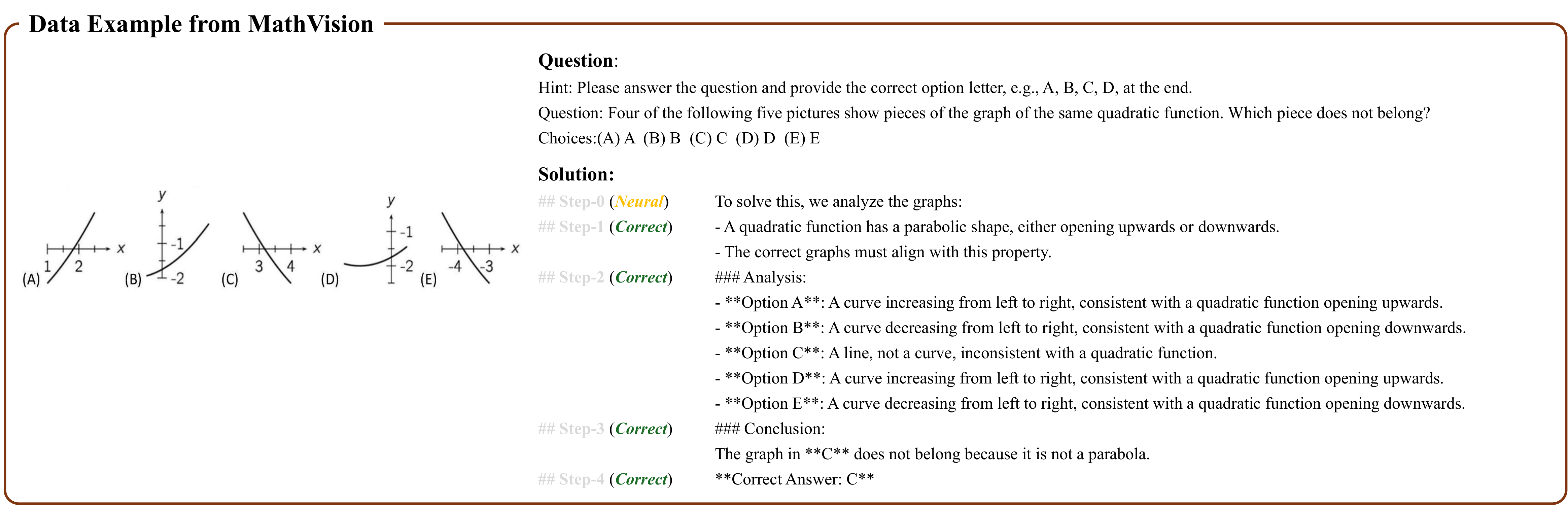}}
    \caption{}
\end{subfigure}
\begin{subfigure}[b]{\textwidth}
    {\includegraphics[width=\textwidth]{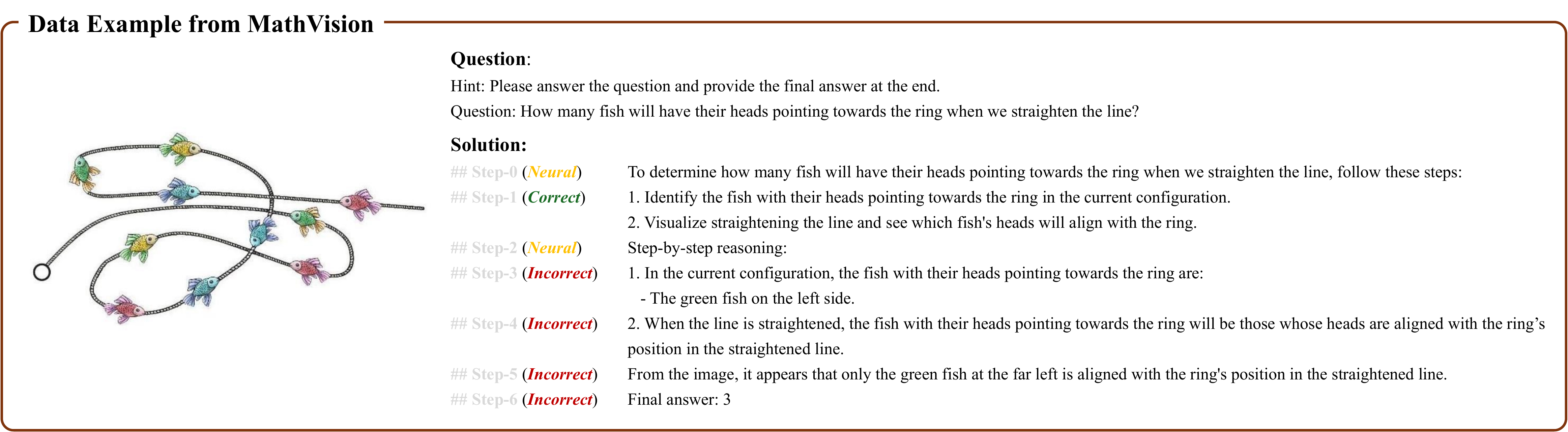}}
    \caption{}
\end{subfigure}
\begin{subfigure}[b]{\textwidth}
    {\includegraphics[width=\textwidth]{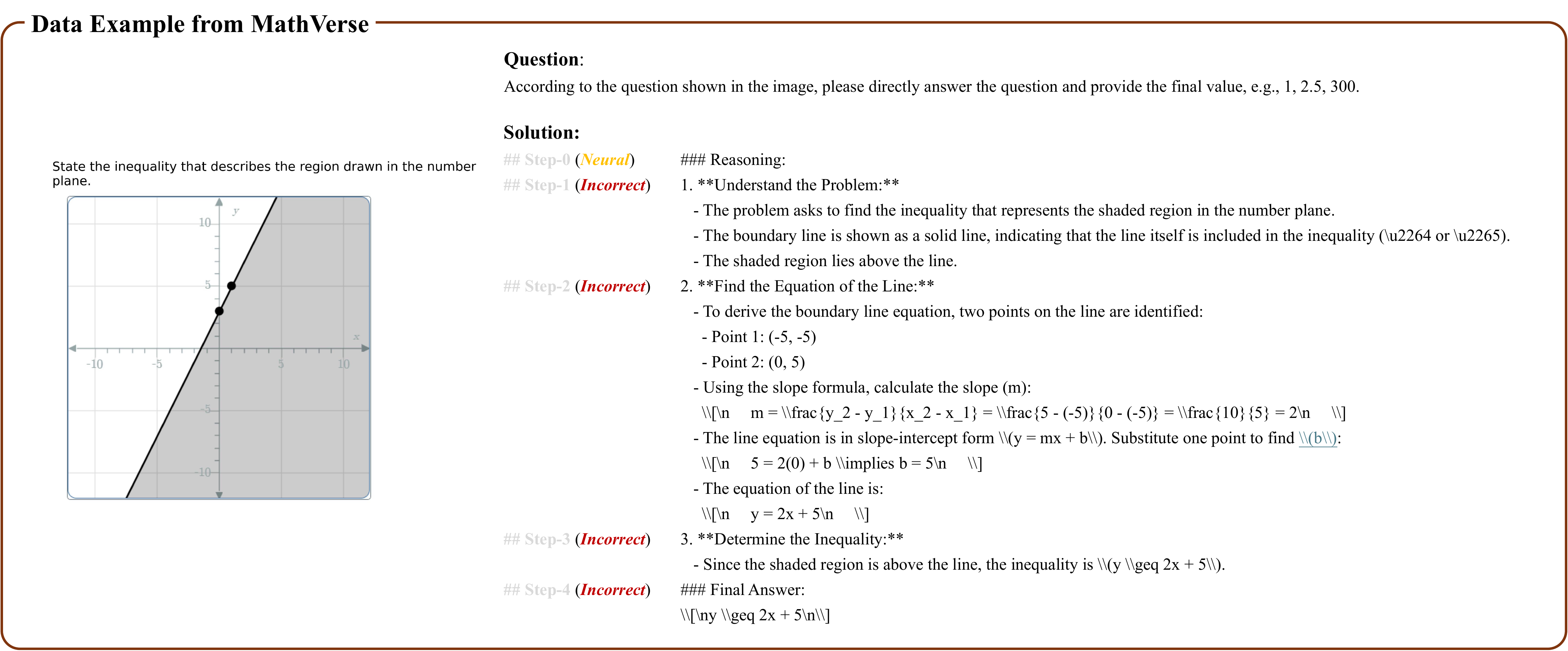}}
    \caption{}
\end{subfigure}
\end{figure*}

\addtocounter{figure}{-1}

\begin{figure*}[t]
\centering
\begin{subfigure}[b]{\textwidth}
    \setcounter{subfigure}{7} %
    {\includegraphics[width=\textwidth]{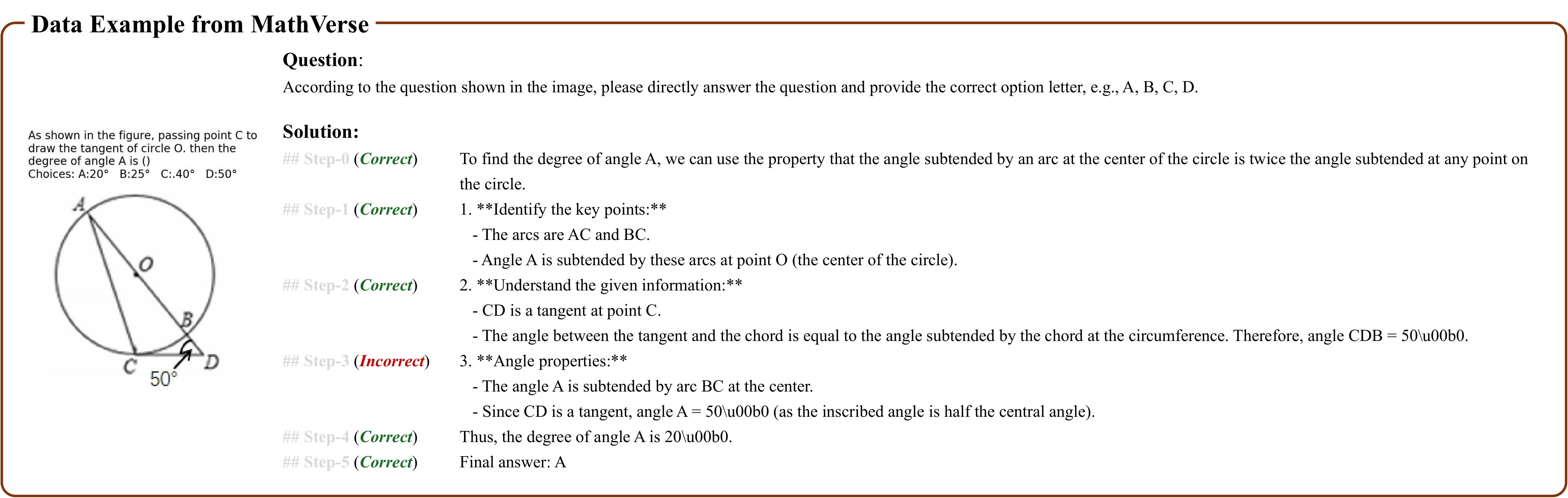}}
    \caption{}
\end{subfigure}
\begin{subfigure}[b]{\textwidth}
    {\includegraphics[width=\textwidth]{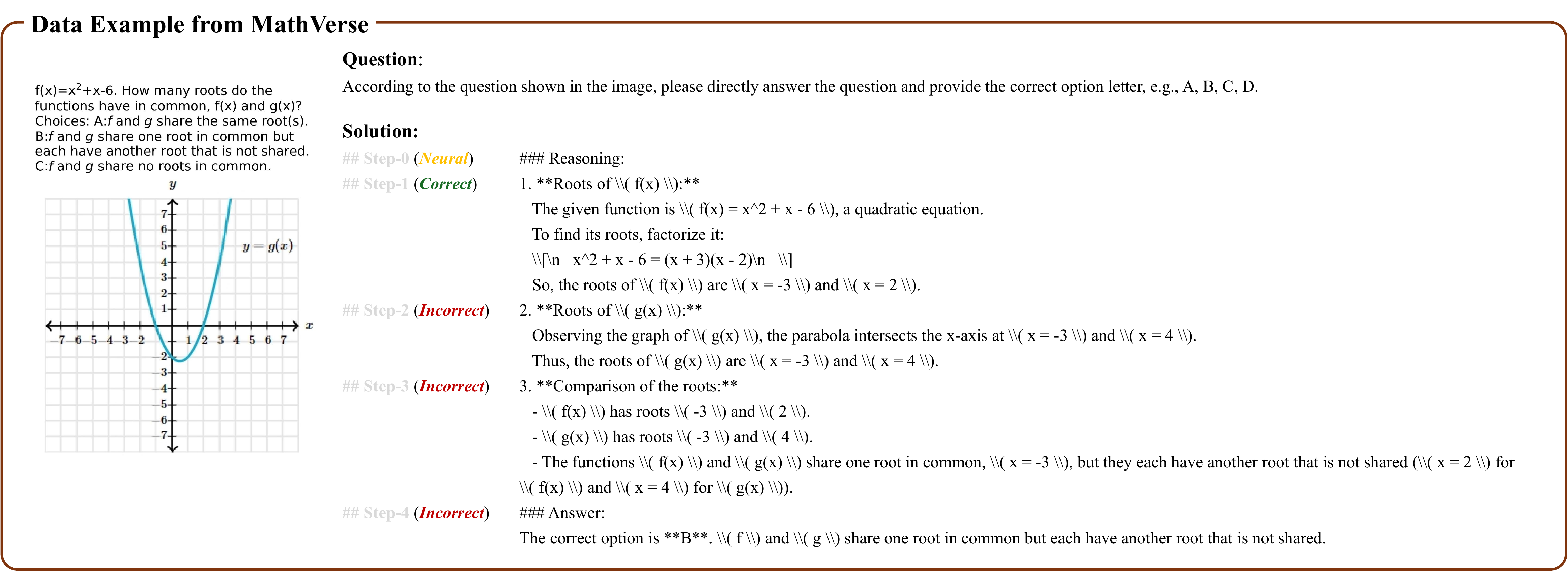}}
    \caption{}
\end{subfigure}
\begin{subfigure}[b]{\textwidth}
    {\includegraphics[width=\textwidth]{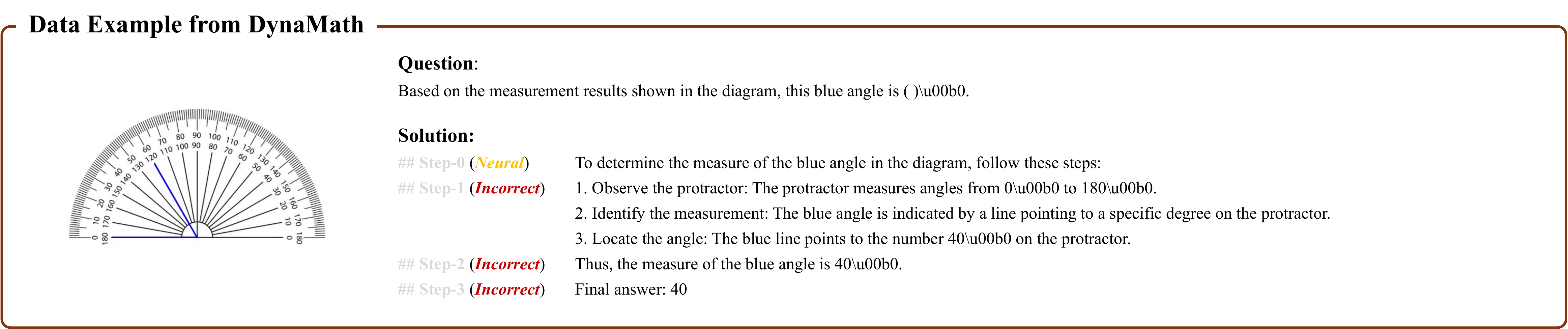}}
    \caption{}
\end{subfigure}
\begin{subfigure}[b]{\textwidth}
    {\includegraphics[width=\textwidth]{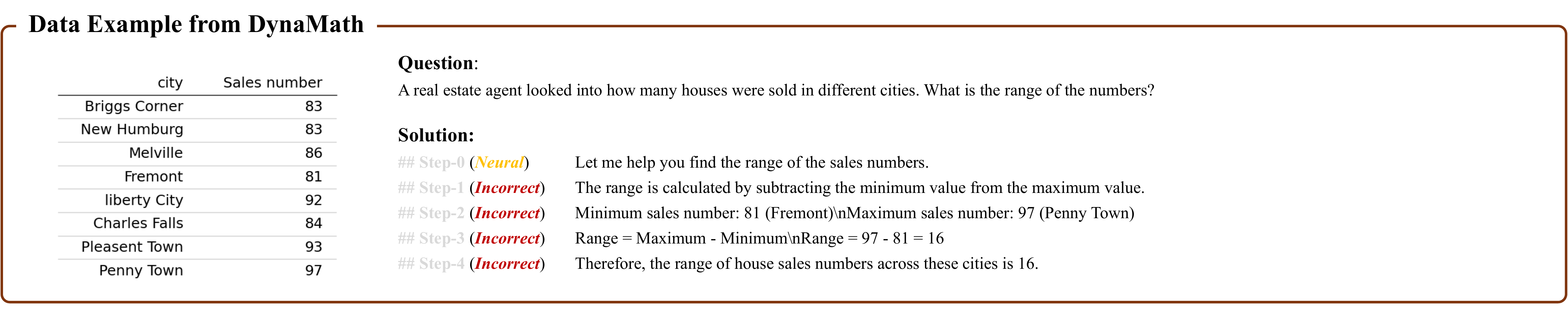}}
    \caption{}
\end{subfigure}
\end{figure*}

\addtocounter{figure}{-1}

\begin{figure*}[t]
\centering
\begin{subfigure}[b]{\textwidth}
    \setcounter{subfigure}{11} %
    {\includegraphics[width=\textwidth]{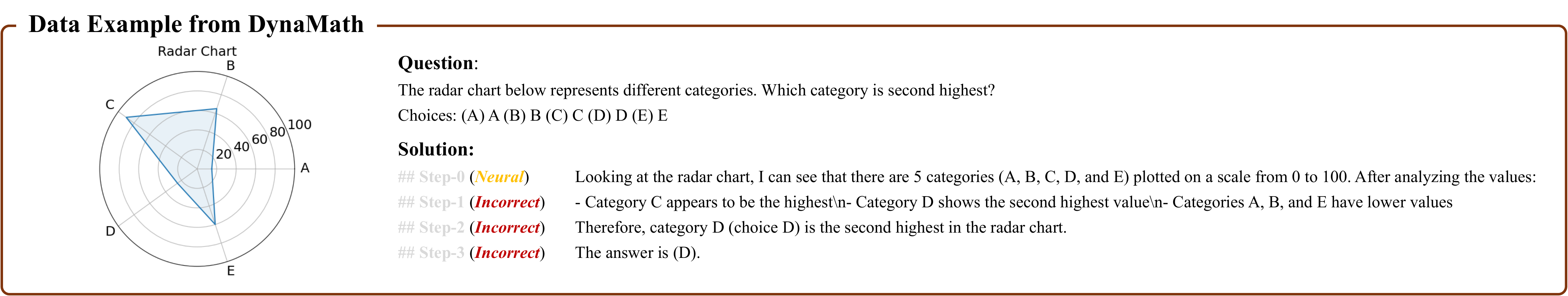}}
    \caption{}
\end{subfigure}
\begin{subfigure}[b]{\textwidth}
    {\includegraphics[width=\textwidth]{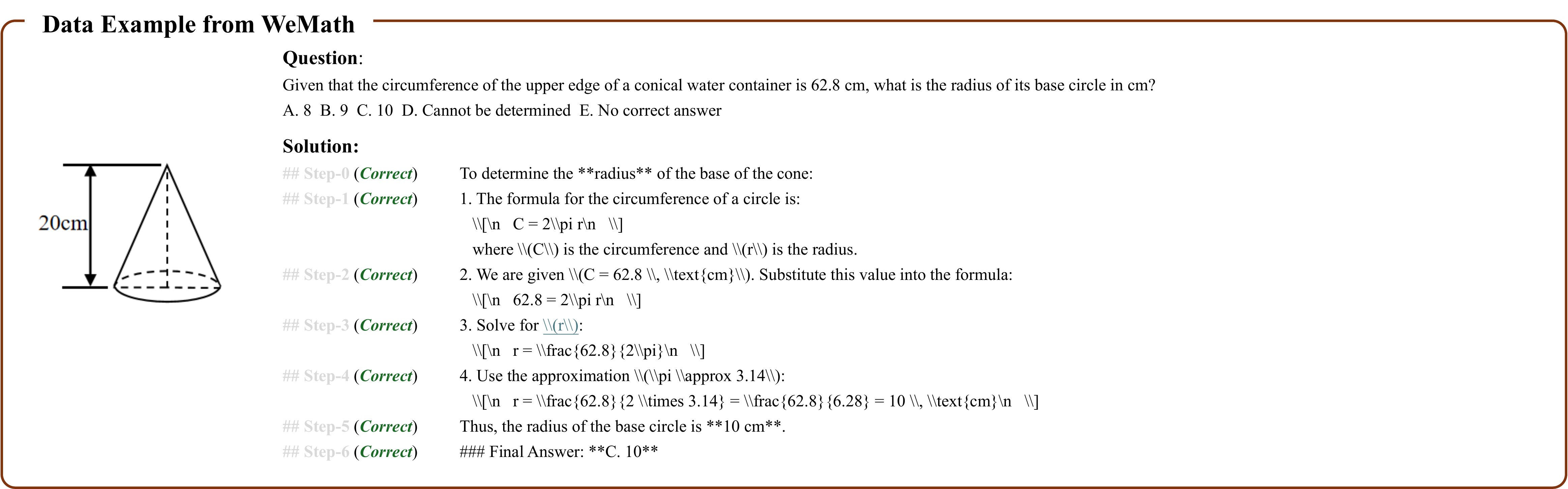}}
    \caption{}
\end{subfigure}
\begin{subfigure}[b]{\textwidth}
    {\includegraphics[width=\textwidth]{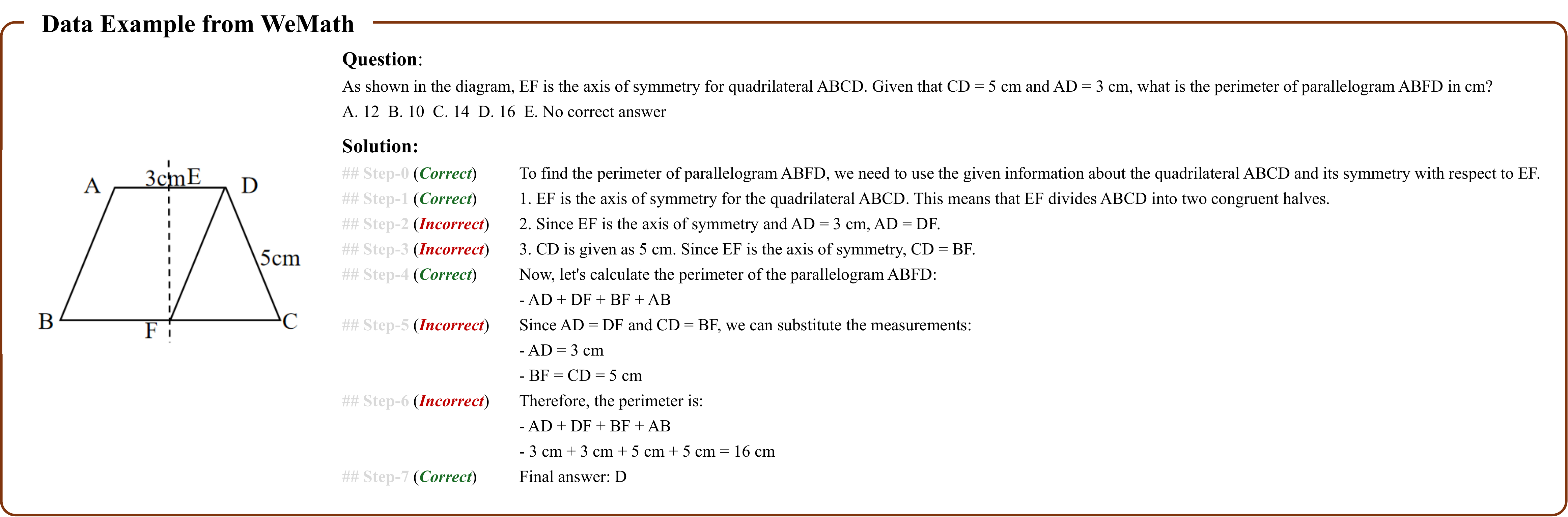}}
    \caption{}
\end{subfigure}
\begin{subfigure}[b]{\textwidth}
    {\includegraphics[width=\textwidth]{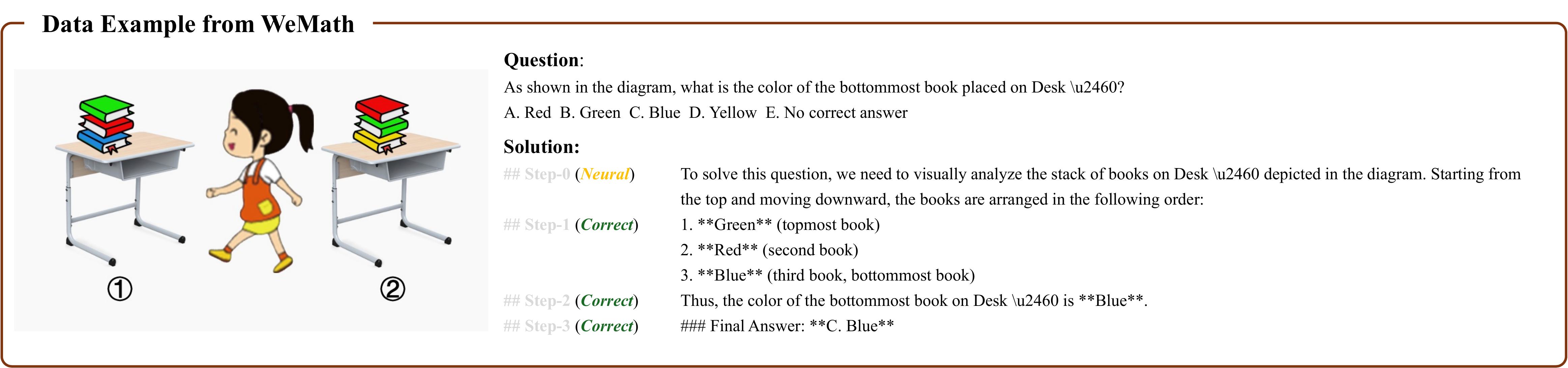}}
    \caption{}
\end{subfigure}
\caption{
\textbf{More data examples from {\benchmarkname}.}
}
\label{fig:suppl-benchmark-examples}
\end{figure*}

\begin{figure*}[t]
\centering
{\includegraphics[width=\textwidth]{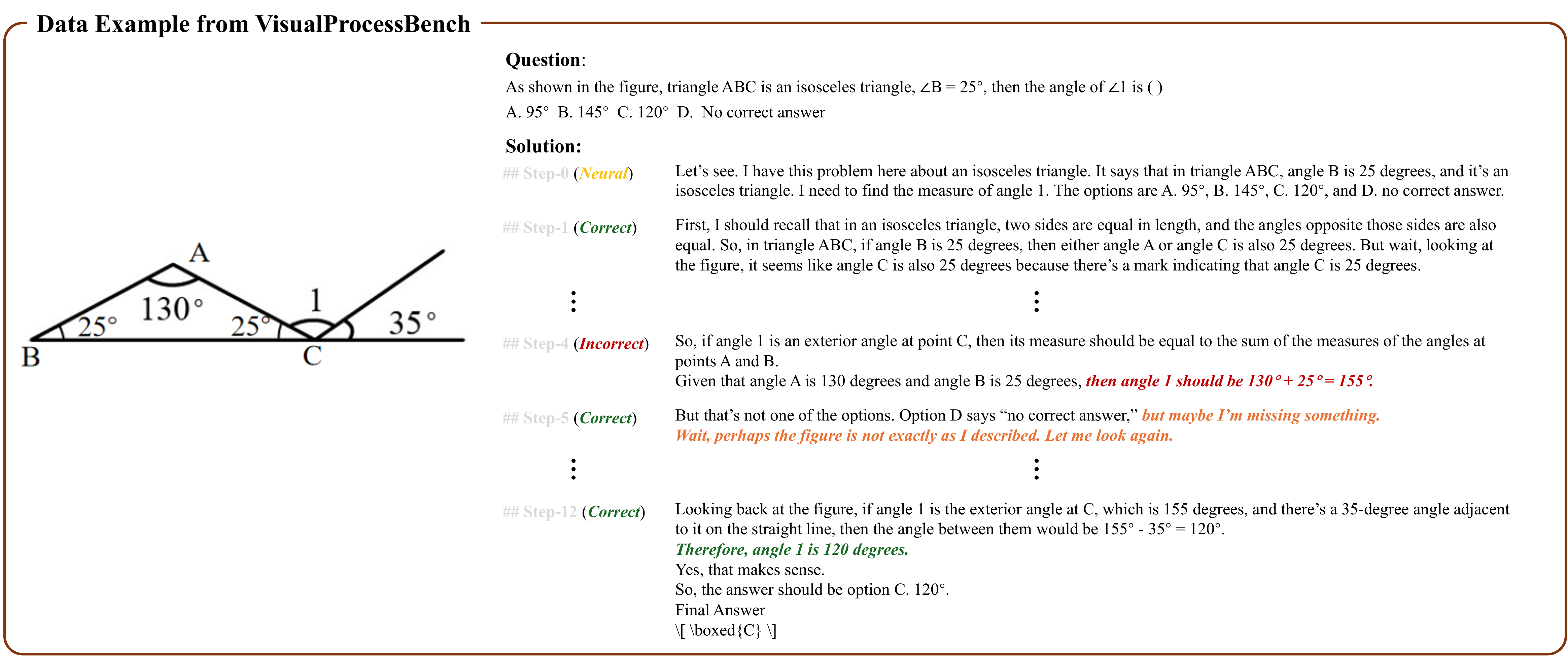}}
\caption{
\textbf{Data example with model reflection from {\benchmarkname}.}
\textcolor{red}{Red} highlights the incorrect answer, \textcolor{orange}{orange} highlights the reflection words, and \textcolor{mygreen}{green} highlights the correct answer.
}
\label{fig:suppl-benchmark-reflection}
\end{figure*}

\end{document}